\documentclass{article} % For LaTeX2e
\usepackage{iclr2027_conference,times}

\usepackage{amsmath,amsfonts,bm}

\def\eqref#1{equation~\ref{#1}}
\def\1{\bm{1}}

\def\ve{{\bm{e}}}

\def\vx{{\bm{x}}}

\def\vz{{\bm{z}}}

\DeclareMathAlphabet{\mathsfit}{\encodingdefault}{\sfdefault}{m}{sl}
\SetMathAlphabet{\mathsfit}{bold}{\encodingdefault}{\sfdefault}{bx}{n}

\newcommand{\R}{\mathbb{R}}

\usepackage{booktabs,graphicx,array}
 
\usepackage{longtable}  

\usepackage{hyperref}
\usepackage{url}
\usepackage[margin=1in]{geometry}
\usepackage{amsmath,amssymb,amsthm}
\usepackage{bm}
\usepackage{mathtools}
\usepackage{graphicx}
\usepackage{booktabs}
\usepackage{hyperref}
\usepackage{booktabs}
\usepackage{multirow}

\newcommand{\vphi}{\bm{\varphi}}
\newcommand{\vrho}{\bm{\rho}}
\newcommand{\norm}[1]{\left\lVert #1 \right\rVert}

\title{AdaKerNet: Neural Kernel Decoding for Task-Adaptive Prediction with Multimodal Large Models}

\author{
Konstantinos D. Polyzos$^{1,*}$,
Eleni Oikonomou$^{*}$ \&
Tara Javidi$^{1}$ \\
$^{1}$University of California San Diego \\
$^{*}$Equal contribution.
}

\iclrfinalcopy % Display authors and disable review line numbers.
\begin{document}

\maketitle
\fancyhead{} % No conference or review header in the arXiv version.
\renewcommand{\headrulewidth}{0pt}

\begin{abstract}

Large foundation models have been introduced with the promise of efficient adaptation to downstream tasks. Yet, under limited supervision, multimodal large language models (MLLMs), an important class of large foundation models, remain challenging to adapt to various downstream tasks. Adaptation typically relies either on MLLM parameter fine-tuning or on training neural-based decoders. Both approaches struggle under limited supervision, while fine-tuning additionally requires access to model parameters, which is often unavailable for closed-source models. We introduce AdaKerNet, a novel learnable task-adaptive neural kernel decoder. AdaKerNet is fully agnostic to the
parameters of the underlying MLLM and operates solely on its (frozen) rich representations obtained from the diverse available modalities. AdaKerNet relies on (i) a set of learnable, Lipschitz-controlled multimodal features derived from these MLLM representations; (ii) a reference kernel that provides a soft structural prior on those features; and (iii) a lightweight nonlinear neural predictor that \textit{adaptively deforms} that structure. Learning the kernel representation and the neural predictor \textit{jointly} within a unified optimization framework allows AdaKerNet to capture features and geometric relationships relevant to the downstream task. Numerical tests across four MLLMs: BLIP-2, LLaVA-1.5, Qwen2.5-VL, and Gemini Embedding 2, and multimodal inputs spanning text, audio, images, and tabular measurements demonstrate significant and consistent improvements over direct MLP,  attention-, autoencoder- and kernel-based decoders, across a range of scarce-label budgets, with average error reduction of up to $41\%$ across baselines based on the $R^2$ metric on continuous prediction tasks. These results establish task-adaptive neural kernel decoding as an effective approach for prediction from frozen multimodal representations in the scarce label regime. Additional structural ablations highlight the complementary contributions of AdaKerNet’s three components, particularly under limited supervision.

\end{abstract}

\section{Introduction}

Foundation models were introduced with the promise of enabling efficient adaptation to a broad range of downstream tasks \cite{Bommasani_etal2021}. However, adapting large multimodal language models (MLLMs), a prominent class of modern foundation models, to downstream prediction tasks remains challenging, particularly under limited supervision. This has meant that existing adaptation strategies typically rely on fine-tuning the MLLM itself, including parameter-efficient variants.  However, the effectiveness of such approaches can markedly degrade in settings where task-specific supervision is limited \cite{Huang_etal2025, Mrabah_etal2025, Farina_etal2025}; furthermore fine-tuning additionally requires access to internal model parameters, which are unavailable for proprietary models exposed only through inference or embedding APIs. 
These limitations motivate an alternative paradigm aligned with the original promise of foundation models: keeping the pretrained MLLM frozen as a representation extractor and concentrating task adaptation entirely within a downstream decoder.
This is the central question of our paper: \textit{What type of decoder can effectively translate frozen MLLM representations into accurate downstream predictions especially when task supervision is scarce?}

 While a reference kernel can provide useful structure for prediction, the predictive performance of vanilla kernel decoding though depends strongly on whether the chosen kernel captures similarities relevant to the downstream target task. Identifying such a kernel is challenging for rich multimodal representations, particularly when limited supervision provides little guidance for kernel selection.
 A natural alternative is to learn a decoder with a simple structure that adapts these representations through task-specific supervision.
Natural choices include multilayer perceptrons (MLPs), autoencoders, transformers and other attention-based neural architectures; their general flexibility allows for learning nonlinear mappings from embeddings to target variables with attention-based modules additionally modeling interactions among available representation components or tokens. Although flexible, these neural architectures must infer the relevant predictive structure from the available supervision and often fall short in sparse data regimes with limited supervision; see e.g., \cite{battaglia2018relational}.

To this end, we propose AdaKerNet, a novel neural kernel decoding architecture that unifies kernel representation learning and neural prediction within a joint optimization framework. The framework consists of (i) a \textit{Lipschitz-controlled feature extractor}  which transforms MLLM representations into a learnable (often lower dimensional) latent representation whose geometry is controlled to align with a given kernel; and a subsequent step of (ii) \textit{Task-adaptive kernel deformation} where the decoder learns an explicit deformation/correction to the kernel feature map via a (shallow) MLP prediction head. Specifically, kernel reconstruction and supervised prediction jointly shape the induced feature map, allowing task adaptation while regularizing deviations from a reference kernel. The learned kernel remains positive semidefinite by construction, while its explicit learned features serve as the basis of a jointly learned nonlinear neural predictor.

\begin{figure}[t]
    \centering
    \IfFileExists{AdaKerNet_overview.pdf}{%
        \includegraphics[width=0.95\linewidth]{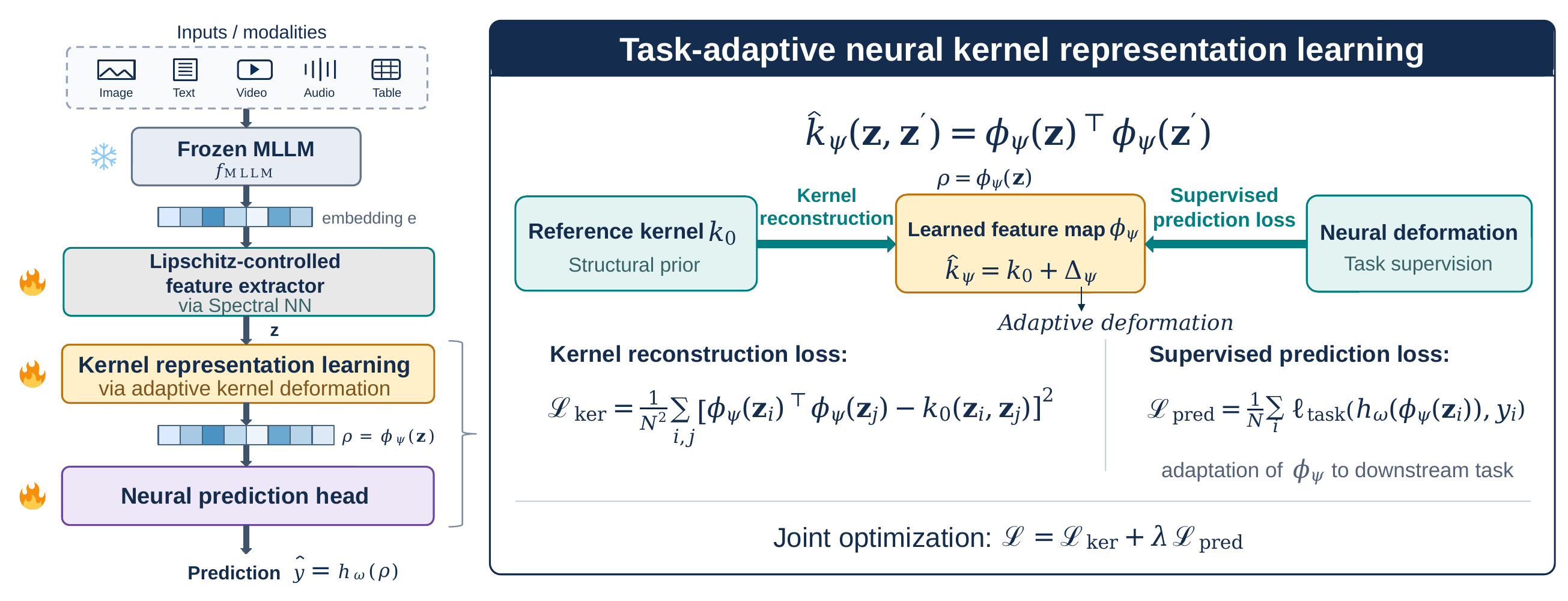}%
    }{%
        \fbox{\parbox[c][3.2cm][c]{0.84\linewidth}{\centering
        Replace this placeholder with
        \texttt{AdaKerNet\_overview.pdf}.}}%
    }
    \caption{ \textbf{Overview of AdaKerNet}: (1) A spectrally normalized neural network-based feature extractor transforms frozen MLLM embeddings into Lipschitz-controlled features, controlling the geometry of the subsequent decoding stage(s), and facilitating alignment with a reference kernel that encodes a suitable structural prior; (2) A task-adaptive neural kernel decoder that extracts a nonlinear neural kernel feature map to deform and adjust the prior selection of reference kernel to the task-specific supervision data; and (3) an MLP prediction head that maps the learned kernel feature to the output labels. Kernel reconstruction anchors the induced feature map to the reference kernel, while supervised prediction adaptively deforms it to capture task-relevant relationships. The adaptively learned kernel feature map serves as the basis of the jointly trained nonlinear neural (MLP) predictor.}
    \label{fig:architecture}
\end{figure}

Through careful empirical evaluations and ablation studies, we show that the proposed approach significantly outperforms both vanilla kernel decoding as well as other deep learning approaches including MLPs, attention-based transformers, and auto-encoders in continuous prediction downstream tasks. More specifically, our contributions can be contextualized as follows: 

\begin{enumerate}
    \item \textit{A unified formulation of kernel structure and neural prediction}. Leveraging a generic and spectrally normalized (shallow) neural network for Lipschitz-controlled feature extraction, the MLLM embedding is transformed to features whose geometry is aligned with the reference kernel acting as a suitable structural prior. Building on these features, we jointly learn a kernel feature map and a nonlinear neural predictor, formulating kernel representation learning as a task-adaptive deformation/correction of the reference kernel. A kernel reconstruction penalty regulates this deformation, balancing fidelity to the reference similarities with supervised adaptation to the downstream task.
    \item \textit{An MLLM-agnostic decoding architecture}. Our framework operates on frozen representations without modifying the underlying MLLM encoder, making it applicable to both accessible pretrained models and proprietary closed-source embedding services, as well as heterogeneous combinations of diverse modalities including text, audio, images, and tabular measurements.
    \item \textit{Empirical evaluation under limited supervision}. We evaluate the proposed decoder across multiple MLLM representations and label budgets in continuous prediction tasks, with comparisons against direct conventional MLP, autoencoders and attention-based decoders, along with kernelized methods including fixed-kernel ridge regression, and spectral neural Gaussian process predictors. The reported results demonstrate consistent improvements over direct neural decoders and gains over kernel-based counterparts in almost all evaluated settings and (sparse) labeled data budgets.
\end{enumerate}

\section{Related Work}
\label{sec:related_work}

\paragraph{Efficient fine-tuning adaptation of multimodal large language models.}
Parameter-efficient fine-tuning provides an alternative to updating all parameters of a pretrained large model. LoRA~\citep{hu2022lora} represents weight updates through trainable low-rank matrices while keeping the original weights frozen, substantially reducing the number of optimized parameters. QLoRA~\citep{dettmers2023qlora} further reduces memory requirements by backpropagating through a frozen, four-bit quantized model into low-rank adapters. These approaches make adaptation more computationally accessible, although their original formulations and evaluations primarily concern language models. In the multimodal setting, LLaMA-Adapter~\citep{zhang2023llamaadapter} introduces learnable adaptation prompts and gated attention, while LLaMA-Adapter V2~\citep{gao2023llamaadapterv2} extends this approach with additional trainable parameters and visual instruction learning. These methods demonstrate that substantial multimodal adaptation can be achieved while updating only a small fraction of the model. 

Nevertheless, computational efficiency does not guarantee statistical efficiency when available supervision is very limited; although these methods reduce the number of trainable parameters,
they still rely on labeled examples to learn task-specific updates,
which can be challenging to estimate reliably in the very sparse-label
regime. We therefore focus on the naturally complementary setting in which the pretrained large model/ representation extractor remains fixed and learning is confined to a downstream decoder. This setting is also relevant to proprietary MLLMs: when a model exposes only an inference or embedding API, without parameter access or a supported fine-tuning interface, users cannot directly apply LoRA, QLoRA, or internal adapter training. We note that due to space constraints, we discuss only a representative selection of fine-tuning-based approaches, as these are not the main focus of this work. Our approach requires only the extracted representations without access to the underlying model parameters or structure. 

\vspace{-0.2cm}

\paragraph{Prediction from frozen multimodal representations.} Recent studies demonstrate that frozen multimodal representations
contain predictive information that is not fully reflected in
the models' generated textual responses. \citet{mitra2025sav} introduce Sparse Attention Vectors (SAVs),
selecting discriminative internal attention-head activations for
few-shot classification through class-centroid similarities and
voting; yet this approach is designed solely for classification task. \citet{li2026dpcvqa} further show that a frozen MLLM can serve as a perceptual prior for video quality assessment, with only a lightweight calibration module trained on its outputs and hidden states. These findings naturally motivate direct prediction from pretrained multimodal representations. Yet, designing a decoder that balances task adaptation with structural guidance is nontrivial and challenging, particularly under limited supervision.

Relying on frozen MLLM representations, task-specific learning is confined
to a downstream decoder, allowing the extracted embeddings to be
(re)used without updating the underlying MLLM. Existing approaches include linear probes on frozen
features~\citep{radford2021clip},
MLP regression heads, optionally preceded by sequence-processing
modules~\citep{cui2025m3agiqa},
and attention-based modules that model interactions among visual
tokens and textual features~\citep{yu2025rankingaware}.
CLIP-Adapter~\citep{gao2021clipadapter} learns bottleneck transformations
with residual connections to preserve pretrained information while
adapting features for few-shot classification.
Tip-Adapter~\citep{zhang2022tipadapter} incorporates labeled examples
through a key--value cache and similarity-based retrieval, with an
optional fine-tuning stage.

Our method advances this line of work by unifying task-adaptive
kernel representation learning and nonlinear neural prediction within a
jointly trained decoder, with a reference-kernel reconstruction
penalty providing explicit structural guidance under limited supervision.
Whereas a directly supervised MLP or attention-based head must infer
task-relevant relationships from the available labels, our decoder
additionally receives structural guidance from a reference kernel.
The kernel-reconstruction objective supplies a pairwise training
signal that does not require target labels, providing an additional
inductive bias when supervision is scarce.
Unlike preserving pretrained features,
this objective regularizes their induced similarity structure while
allowing the learned features themselves to change.

Crucially, the reference similarities are not imposed as fixed:
joint optimization with the MLP prediction loss allows the learned
kernel to deform toward relationships that better support the
downstream task.
The decoder therefore combines the structural guidance of kernel
representations with the flexibility of nonlinear neural prediction.
This balance is particularly relevant in the sparse-label regime,
where unconstrained adaptation can overfit, while a fixed similarity
function may be poorly aligned with the target.
Furthermore, our decoder operates directly on fixed-length embeddings
without requiring token-level features, learnable input prompts,
modality-specific attention mechanisms, or backbone fine-tuning.
It can consequently be applied across different MLLMs and modality
combinations, including closed-source proprietary services that only reveal embeddings.

\vspace{-0.3cm}

\paragraph{Kernel representation learning and neural prediction.}
Kernel methods encode relationships among samples through positive-semidefinite similarity functions. With a fixed kernel, kernel ridge regression optimizes prediction coefficients while retaining the prescribed similarity structure. Its performance consequently depends on whether the selected kernel captures relationships relevant to the target/downstream task \citep{vovk2013kernel}. Kernel approximation methods address a different concern: random Fourier features~\citep{rahimi2007random} provide explicit finite-dimensional approximations to shift-invariant kernels, enabling efficient learning with linear predictors. Approximating a prescribed kernel, however, does not itself adapt that kernel to the prediction task.

Learning task-dependent kernels has been explored in the literature. Centered-alignment methods~\citep{cortes2012alignment} learn kernels using similarity measures between kernel matrices and include formulations that jointly learn the kernel and predictor. Deep kernel learning~\citep{wilson2016deep} combines neural feature transformations with Gaussian-process covariance functions, jointly learning their parameters through the marginal likelihood. A complementary approach uses ensembles of Gaussian processes
with different candidate kernels, adaptively updating their
model weights as observations become available to reduce reliance
on a single preselected kernel ~\citep{lu2023surrogate}. Tuned random features (TRF)~\citep{shilton2022trf} learns translation-invariant kernels through their spectral densities, reducing kernel selection to regularized risk minimization within a random-feature formulation. These approaches already address limitations of fixed similarities; our work builds on this broader aspect through a different complementary perspective: \textit{Once a suitable (possibly learned) kernel is selected as a structural prior/reference kernel,
can it be further deformed/corrected adaptively to better align with the downstream task, combining the merits of kernel representation learning and nonlinear
neural prediction?}

One of our main contributions lies on a specific formulation that combines reference-kernel reconstruction with nonlinear neural prediction on frozen multimodal representations. We learn an explicit feature map whose inner products define a positive-semidefinite kernel and regularize its Gram/kernel matrix toward that of a reference kernel. Joint training with an MLP prediction head permits supervised deviations from this reference structure. The formulation therefore addresses fixed-kernel mismatch while retaining a structural constraint absent from an otherwise unconstrained neural decoder. Unlike the spectral formulation of TRF, the learned feature map need not induce a translation-invariant kernel; unlike conventional deep kernel learning, prediction is performed by an MLP on explicit kernel features rather than a Gaussian-process model. This provides a unified perspective on kernel representation learning and neural MLP prediction: the reference kernel supplies a soft similarity prior, the feature map learns its task-adaptive deformation, and the neural head maps the resulting features to downstream targets.

\vspace{-0.3cm}

\section{Method}

\vspace{-0.2cm}

We propose AdaKerNet, a neural kernel decoding framework for prediction with MLLMs, illustrated in Fig.~\ref{fig:architecture}, operating on (1) a frozen MLLM and its output embedding $e$; (2) a Lipschitz-controlled feature extraction block; (3) a task-adaptive kernel deformation block; and (4) an MLP prediction head that is jointly trained as part of the preceding task-adaptive optimization stage. Specifically,

\textbf{3.1. Multimodal Encoder}. Let $\vx=(\vx^{(1)},\dots,\vx^{(M)})$ denote an input observed through
$M$ modalities, and let $y$ denote its associated target/label. A pretrained MLLM $f_{\mathrm{MLLM}}$ maps the multimodal input to a
$d$-dimensional embedding,
\begin{equation}
    \ve
    =f_{\mathrm{MLLM}}(\vx^{(1)},\dots,\vx^{(M)})
    \in\R^d.
    \label{eq:mllm-embedding}
\end{equation}
In the scarce label regime where LoRA and similar approaches struggle, or when MLLMs are closed-source making such approaches impractical, the alternative option is to keep $f_{\mathrm{MLLM}}$ frozen and use $\ve$ as a fixed-length
summary of the fused multimodal content and subsequently work on top of these for the downstream prediction task.

\textbf{3.2. Lipschitz-controlled Features Block}. This block maps the MLLM embedding to a latent representation
\begin{equation}
    \vz=g_\theta(\ve)\in\R^D.
    \label{eq:block-a}
\end{equation}
The aim is to extract representations that allow for the 
(Lipschitz-controlling) property
\begin{equation}
    \norm{g_\theta(\ve) - g_\theta(\ve')}_2
    \;\le\;
    \beta\,\norm{\ve - \ve'}_2,
\end{equation}
meaning that semantic distances in embedding space are not
exploded.

In this work, we consider $g_\theta$ to be an $L$-layer neural network with weight matrices
$\{W_\ell\}_{\ell=1}^L$ and $1$-Lipschitz activations. Each weight matrix
is spectrally normalized as
\begin{equation}
    \widetilde W_\ell
    =c_\ell\frac{W_\ell}{\sigma_{\max}(W_\ell)},
    \qquad
    \sigma_{\max}(W_\ell)
    =\max_{\norm{\bm{u}}_2=1}\norm{W_\ell\bm{u}}_2,
    \label{eq:spectral-normalization}
\end{equation}
where $c_\ell>0$ controls the Lipschitz constant of layer $\ell$, and the spectral norm
$\sigma_{\max}(W_\ell)$ can be estimated efficiently by power iteration.  This
constrains the expansion of distances according to
\begin{equation}
    \norm{g_\theta(\ve)-g_\theta(\ve')}_2
    \leq
    \left(\prod_{\ell=1}^L c_\ell\right)
    \norm{\ve-\ve'}_2.
    \label{eq:lipschitz-bound}
\end{equation}

To align $\vz$ with the structural prior induced by a reference kernel while simultaneously supervising the feature extractor for the downstream prediction task, we apply a kernel ridge regression head to $\vz$ using the reference kernel to map $\vz$ to  $y$. In this way, the learned representation is shaped such that its geometry (i) remains Lipschitz-controlled and (ii) is compatible with the chosen reference kernel. Once the learning of $\vz$ is complete, it is subsequently passed to the next stage.

\textbf{3.3. Task-adaptive Kernel Deformation Block}.
Let $k_0:\R^D\times\R^D\rightarrow\R$
denote a fixed positive-semidefinite \emph{reference kernel}. This kernel
encodes a prior similarity structure in the latent space, but we do \textit{not}
assume that it is perfectly aligned with the downstream task.

This block learns a nonlinear feature map
\begin{equation}
    \vphi_\psi:\R^D\rightarrow\R^F,
    \qquad
    \vrho=\vphi_\psi(\vz),
    \label{eq:kernel-embedding}
\end{equation}
which induces the learned kernel
\begin{equation}
    \widehat{k}_\psi(\vz,\vz')
    :=
    \vphi_\psi(\vz)^{\!\top}\vphi_\psi(\vz')
    .
    \label{eq:learned-kernel}
\end{equation}
By construction, $\widehat{k}_\psi$ is positive semidefinite for every
choice of $\psi$. Indeed, for arbitrary points
$\{\vz_i\}_{i=1}^N$ and coefficients $\{a_i\}_{i=1}^N$,
\begin{equation*}
    \sum_{i=1}^N\sum_{j=1}^N
    a_i a_j\widehat{k}_\psi(\vz_i,\vz_j)
    =
    \norm{\sum_{i=1}^N a_i\vphi_\psi(\vz_i)}_2^2
    \geq 0.
    \label{eq:learned-kernel-psd}
\end{equation*}

To expose the adaptation explicitly, we express
\begin{equation}
    \widehat{k}_\psi(\vz,\vz')
    =
    k_0(\vz,\vz')+\Delta_\psi(\vz,\vz')
    ,
    \label{eq:kernel-deviation}
\end{equation}
where
$
    \Delta_\psi(\vz,\vz')
    :=
    \vphi_\psi(\vz)^{\!\top}\vphi_\psi(\vz')
    -k_0(\vz,\vz')
    \label{eq:deviation-definition}
$
is the learned deviation from the reference kernel. Although
$\widehat{k}_\psi$ is a valid kernel, $\Delta_\psi$ need not itself be a
positive-semidefinite kernel. The kernel loss introduced below controls
the magnitude of this deviation on the observed data.

\textbf{3.4. MLP Prediction Head}.
The kernel representation is passed to an MLP
$h_\omega:\R^F\rightarrow\R$ to predict the scalar (regression) target:
\begin{equation}
    \widehat y_i
    =h_\omega(\vrho_i)
    =h_\omega\!\left(\vphi_\psi(\vz_i)\right)
    .
    \label{eq:mlp-prediction}
\end{equation}

\textbf{Joint training objective}. Let $\mathcal B$ be a minibatch of size $N=|\mathcal B|$. We \underline{jointly}
learn the feature map $\vphi_\psi$ and the prediction head $h_\omega$ by
combining supervised prediction with soft regularization toward the
reference kernel:
\begin{equation}
\begin{aligned}
    \mathcal L(\psi,\omega)
    ={}
    \lambda\underbrace{
    \frac{1}{N}\sum_{i\in\mathcal B}
    \ell_{\mathrm{task}}\!\left(
       h_\omega(\vphi_\psi(\vz_i)),y_i
    \right)
    }_{\mathcal L_{\mathrm{pred}}}
    +
    \underbrace{
    \frac{1}{N^2}\sum_{i,j\in\mathcal B}
    \left[
       \widehat{k}_\psi(\vz_i,\vz_j)-k_0(\vz_i,\vz_j)
    \right]^2
    }_{\mathcal L_{\mathrm{ker}}}.
    \label{eq:general-objective}
\end{aligned}
\end{equation}
Here, $\lambda\geq0$ determines how strongly the learned kernel is adaptively deformed from the reference kernel structure to fit to the downstream task.

For regression task with squared-error loss, the fully expanded objective is
\begin{equation}
\begin{aligned}
    \mathcal L(\psi,\omega)
    ={}
    \lambda\frac{1}{N}\sum_{i\in\mathcal B}
    \left[
       h_\omega\!\left(\vphi_\psi(\vz_i)\right)-y_i
    \right]^2 +
    \frac{1}{N^2}\sum_{i,j\in\mathcal B}
    \left[
       \vphi_\psi(\vz_i)^{\!\top}\vphi_\psi(\vz_j)
       -k_0(\vz_i,\vz_j)
    \right]^2.
\end{aligned}
\label{eq:loss-scalar}
\end{equation}

We interpret the learning of $\widehat{k}_\psi$ as a \emph{task-adaptive deformation} of the reference kernel $k_0$. Intuitively, the
prediction loss can deform the kernel toward a task-suitable similarity structure, while the reference-kernel penalty prevents unconstrained
deviations from the prior reference geometry. In the supplementary material, we provide a constrained optimization interpretation of our approach along with remarks on why the learned kernel is task-adaptive and its relation to kernel ridge regression. 

\vspace{-0.2cm}

\section{Numerical Tests}

\subsection{Experimental Setting}

We evaluate AdaKerNet on six real-world multimodal benchmarks; namely (1) PAD-UFES-20-AGE \cite{pacheco2020padufes20}, (2) PAD-UFES-20-MOLE-LESION \cite{pacheco2020padufes20}, (3) AMAZON-FASHION \cite{hou2024bridging}, (4) QUECHUA-VALENCE \cite{paccotacya2022quechua}, (5) SPEECHOCEAN762-FLUENCY \cite{zhang2021speechocean762} and (6) SPEECHOCEAN762-PROSODIC  \cite{zhang2021speechocean762}, spanning healthcare, e-commerce, affective computing, and language education. To demonstrate the applicability of AdaKerNet across diverse modalities, benchmarks (1)–(3) comprise text, image, and tabular data, whereas benchmarks (4)–(6) comprise audio and text data.  All tasks are formulated as continuous prediction (regression) problems; further details are provided in the supplementary material. As a figure of merit, we evaluate prediction performance on a held-out test set using the coefficient of determination, $R^2$ (higher is better) averaged across 10 independent runs, as it is scale-normalized and therefore more comparable across datasets than absolute-error metrics such as MAE.  To avoid ambiguity, we note that $R^2$ can take negative values, indicating predictive performance worse than simply predicting the mean of the target variable.

\vspace{-0.1cm}
\textit{Frozen MLLM Representation.}
We use frozen pretrained MLLMs as feature extractors. For benchmarks (1)–(3), BLIP-2, LLaVA-1.5, and Qwen2.5-VL are used (separately) to extract multimodal representations, while the closed-source Gemini Embedding 2 model is used for benchmarks (4)–(6). Further details on the representation extraction procedure are provided in the supplementary material.

\vspace{-0.1cm}
\textit{Sparse-label regime.}
In this work, we focus on the predictive performance of MLLM decoders under limited supervision. Accordingly, for each dataset, we evaluate performance across labeled training-set sizes of $\{100,200,300,500,1000\}$ samples. 

\vspace{-0.1cm}
\textit{Baselines.} We compare AdaKerNet with the standard MLLM neural decoder counterparts working directly on the (M)LLM representation $\ve$ (B-I -- B-II); see e.g. \cite{leach2026bounded, tang2024understanding, wen2026beyond, su2026relish}. We additionally consider a suite of less commonly used MLLM decoders that operate directly on the representation~$\ve$ (B-III -- B-V). Although these architectures have not been widely explored for this setting in the literature, they constitute natural and practically relevant baseline choices.
\vspace{-0.25cm}
\begin{itemize}
    \item[\textbf{B-I}] \textbf{MLP Decoder} We use a three layer MLP with 512, 256 and 1 units with ReLU activations in the first two, constituting the optimized architectural design following a thorough architectural search. 
    \item[\textbf{B-II}] \textbf{Transformer Decoder.} We use a two-layer transformer architecture ($d_{\mathrm{model}}=128$, 4 attention heads, feedforward width 512) applied to a token sequence constructed from $\ve$, followed by mean pooling and a linear readout. 
    \item[\textbf{B-III}] \textbf{Autoencoder and Linear Head.}  We use a generic autoencoder whose encoder is a three-layer MLP with 512, 256, and 128 hidden units and
ReLU activations, together with a mirrored decoder. The resulting low-dimensional latent representation, corresponding to the encoder output, is then passed to a linear prediction head trained using the available labeled samples.
    \item[\textbf{B-IV}] \textbf{Kernel Ridge Regression Head}. We use a KRR head with an RBF kernel, with all kernel and regularization hyperparameters selected using the training data.
    \item[\textbf{B-V}] \textbf{Spectral-normalized Neural Gaussian Process Head}. We use an SNGP head introduced in \cite{liu2020simple} to enable more flexible, data-adaptive similarity modeling than a conventional GP with a fixed kernel. The architecture consists of a spectrally normalized two-layer MLP with 512 and 256 hidden units and GELU activations, followed by a GP output layer relying on RBF kernel and implemented using a Random Fourier Feature approximation of 128 random features.
\end{itemize}

\textit{AdaKerNet Implementation Details}.
For the Lipschitz-controlled feature extraction block, we obtain the Lipschitz-controlled feature representation $\vz$ from $\ve$ using a two-layer spectrally normalized neural network with 512 and 256 hidden units, where the second layer outputs $\vz$. To transform the MLLM representations into Lipschitz-controlled features that can be aligned with the structural prior induced by a reference kernel, while simultaneously providing supervision for the feature extractor, $\vz$ is passed to a KRR head using the reference kernel $k_0$ (which is subsequently adaptively deformed in the next stage), to predict the corresponding target $y$.

For the subsequent neural kernel decoding block, the learnable mapping $\vphi_\psi(\cdot)$ is parameterized by a three-layer MLP with 512, 256, and 128 hidden units, while $h_\omega(\cdot)$ is implemented as a two-layer MLP with 256 and 1 units, respectively. Additional implementation details for all components are provided in the supplementary material.

\vspace{-0.1cm}

\subsection{Numerical Results}

\begin{table*}[!t]
\centering
\definecolor{gaingreen}{RGB}{0,128,55}
\caption{Comparison of AdaKerNet with five baselines operating directly on raw MLLM representations across six continuous prediction benchmarks, varying label budgets $N$, and different MLLMs spanning BLIP-2, LLaVA-1.5, Qwen2.5-VL, and Gemini Embedding 2. Performance is measured by mean test $R^2$, with the best result in \textbf{bold} and the second best \underline{underlined}. Avg.\ err.\ red.\ denotes AdaKerNet’s relative reduction in test MSE, computed as $100(R^2_{\mathrm{ours}}-R^2_{\mathrm{base}})/(1-R^2_{\mathrm{base}})$ and averaged over the five baselines.}
\label{tab:main}
\setlength{\tabcolsep}{3pt}
\renewcommand{\arraystretch}{1.0}
\setlength{\aboverulesep}{0pt}\setlength{\belowrulesep}{0pt}
\setlength{\extrarowheight}{0.5pt}
\resizebox{\textwidth}{!}{%
\begin{tabular}{@{}l*{5}{w{c}{4.2em}}@{\hspace{6pt}}|@{\hspace{6pt}}*{5}{w{c}{4.2em}}@{}}
\toprule
\textbf{Method} & $N{=}100$ & $200$ & $300$ & $500$ & $1000$ & $N{=}100$ & $200$ & $300$ & $500$ & $1000$ \\
\midrule
\rule{0pt}{15pt} & & & \makebox[0pt]{\textbf{PAD-UFES-20-AGE / BLIP-2}} & & & & & \makebox[0pt]{\textbf{PAD-UFES-20-AGE / LLaVA}} & & \\[4pt]
MLP & \underline{0.2586} & \underline{0.2841} & \underline{0.3359} & \underline{0.3729} & \underline{0.4129} & \underline{0.2615} & \underline{0.2925} & \underline{0.3399} & \underline{0.3511} & \underline{0.4082} \\
Transformer & 0.2121 & 0.2409 & 0.3009 & 0.3367 & 0.3757 & 0.1883 & 0.1989 & 0.2648 & 0.2729 & 0.3701 \\
AE $+$ linear & -2.7412 & -0.7753 & -0.1098 & 0.1991 & 0.3462 & -2.6360 & -0.8492 & -0.0904 & 0.1722 & 0.3371 \\
KRR & 0.1338 & 0.2386 & 0.2898 & 0.3467 & 0.4120 & -0.1548 & 0.0680 & 0.1569 & 0.2577 & 0.3521 \\
SNGP & 0.1100 & 0.0894 & 0.1997 & 0.2601 & 0.3325 & 0.0956 & 0.1409 & 0.1412 & 0.1887 & 0.2827 \\
\textbf{AdaKerNet} & \textbf{0.2750} & \textbf{0.3251} & \textbf{0.3579} & \textbf{0.3891} & \textbf{0.4261} & \textbf{0.3045} & \textbf{0.3588} & \textbf{0.3958} & \textbf{0.4061} & \textbf{0.4472} \\
Avg.\ err.\ red. & \textcolor{gaingreen}{\textbf{$+$25.1\%}} & \textcolor{gaingreen}{\textbf{$+$23.2\%}} & \textcolor{gaingreen}{\textbf{$+$16.6\%}} & \textcolor{gaingreen}{\textbf{$+$11.6\%}} & \textcolor{gaingreen}{\textbf{$+$7.8\%}} & \textcolor{gaingreen}{\textbf{$+$32.8\%}} & \textcolor{gaingreen}{\textbf{$+$30.2\%}} & \textcolor{gaingreen}{\textbf{$+$25.8\%}} & \textcolor{gaingreen}{\textbf{$+$20.4\%}} & \textcolor{gaingreen}{\textbf{$+$14.6\%}} \\[2pt]
\midrule
\rule{0pt}{15pt} & & & \makebox[0pt]{\textbf{PAD-UFES-20-AGE / Qwen}} & & & & & \makebox[0pt]{\textbf{PAD-UFES-20-MOLE-LESION / BLIP-2}} & & \\[4pt]
MLP & -0.0157 & \underline{0.0691} & \underline{0.1214} & 0.1150 & 0.1604 & -0.0173 & 0.0065 & 0.1067 & 0.1333 & 0.1856 \\
Transformer & -0.0741 & -0.0344 & -0.0162 & 0.0619 & 0.1699 & -0.0541 & 0.0055 & 0.1037 & 0.1752 & 0.2678 \\
AE $+$ linear & -2.8743 & -1.1566 & -0.1991 & 0.0969 & 0.2710 & -4.6294 & -0.7782 & -0.1012 & 0.1980 & 0.3330 \\
KRR & -0.1291 & 0.0347 & 0.1103 & \underline{0.2012} & \underline{0.2991} & -0.0072 & 0.1275 & 0.1882 & 0.2962 & \underline{0.3659} \\
SNGP & \underline{-0.0074} & 0.0040 & 0.0911 & 0.1527 & 0.1966 & \underline{0.0624} & \underline{0.1691} & \underline{0.1948} & \underline{0.3040} & 0.3178 \\
\textbf{AdaKerNet} & \textbf{0.2771} & \textbf{0.3167} & \textbf{0.3522} & \textbf{0.3690} & \textbf{0.3923} & \textbf{0.1777} & \textbf{0.2291} & \textbf{0.2731} & \textbf{0.3347} & \textbf{0.3716} \\
Avg.\ err.\ red. & \textcolor{gaingreen}{\textbf{$+$41.4\%}} & \textcolor{gaingreen}{\textbf{$+$37.9\%}} & \textcolor{gaingreen}{\textbf{$+$32.9\%}} & \textcolor{gaingreen}{\textbf{$+$27.6\%}} & \textcolor{gaingreen}{\textbf{$+$21.7\%}} & \textcolor{gaingreen}{\textbf{$+$31.4\%}} & \textcolor{gaingreen}{\textbf{$+$24.1\%}} & \textcolor{gaingreen}{\textbf{$+$18.3\%}} & \textcolor{gaingreen}{\textbf{$+$13.9\%}} & \textcolor{gaingreen}{\textbf{$+$10.3\%}} \\[2pt]
\midrule
\rule{0pt}{15pt} & & & \makebox[0pt]{\textbf{PAD-UFES-20-MOLE-LESION / LLaVA}} & & & & & \makebox[0pt]{\textbf{PAD-UFES-20-MOLE-LESION / Qwen}} & & \\[4pt]
MLP & -0.0518 & 0.0550 & 0.0978 & 0.1274 & 0.2134 & -0.0678 & -0.1302 & -0.0379 & -0.0288 & 0.0654 \\
Transformer & -0.2471 & -0.1795 & -0.0421 & -0.0297 & 0.1260 & -0.1593 & -0.1750 & -0.0587 & 0.0175 & 0.1051 \\
AE $+$ linear & -3.2469 & -1.0232 & -0.1570 & 0.1349 & 0.2478 & -3.9287 & -1.0305 & -0.3210 & -0.0194 & 0.1755 \\
KRR & -0.1783 & 0.0425 & \underline{0.1157} & \underline{0.2227} & \underline{0.2971} & -0.0529 & \underline{0.0301} & \underline{0.0699} & \underline{0.1479} & \underline{0.2238} \\
SNGP & \underline{0.0266} & \underline{0.0655} & 0.0997 & 0.1721 & 0.2262 & \underline{-0.0144} & -0.0089 & -0.0080 & 0.0615 & 0.1739 \\
\textbf{AdaKerNet} & \textbf{0.1475} & \textbf{0.2134} & \textbf{0.2497} & \textbf{0.2925} & \textbf{0.3246} & \textbf{0.1356} & \textbf{0.1731} & \textbf{0.1987} & \textbf{0.2380} & \textbf{0.2724} \\
Avg.\ err.\ red. & \textcolor{gaingreen}{\textbf{$+$34.1\%}} & \textcolor{gaingreen}{\textbf{$+$29.0\%}} & \textcolor{gaingreen}{\textbf{$+$22.4\%}} & \textcolor{gaingreen}{\textbf{$+$18.4\%}} & \textcolor{gaingreen}{\textbf{$+$12.7\%}} & \textcolor{gaingreen}{\textbf{$+$31.9\%}} & \textcolor{gaingreen}{\textbf{$+$29.7\%}} & \textcolor{gaingreen}{\textbf{$+$24.2\%}} & \textcolor{gaingreen}{\textbf{$+$20.6\%}} & \textcolor{gaingreen}{\textbf{$+$14.2\%}} \\[2pt]
\midrule
\rule{0pt}{15pt} & & & \makebox[0pt]{\textbf{AMAZON-FASHION / BLIP-2}} & & & & & \makebox[0pt]{\textbf{AMAZON-FASHION / LLaVA}} & & \\[4pt]
MLP & \textbf{0.2152} & \underline{0.2748} & \underline{0.3260} & \underline{0.3651} & \underline{0.4311} & \underline{0.0300} & -0.0009 & 0.1133 & 0.1709 & 0.2505 \\
Transformer & 0.1926 & 0.2111 & 0.2735 & 0.3062 & 0.3883 & 0.0239 & 0.0592 & 0.1361 & 0.1796 & 0.3010 \\
AE $+$ linear & -2.2390 & -0.9056 & -0.1167 & 0.1385 & 0.2850 & -2.9468 & -0.9345 & -0.1608 & 0.1263 & 0.2575 \\
KRR & 0.0154 & 0.1808 & 0.2553 & 0.3326 & 0.4185 & -0.0721 & \underline{0.0868} & \underline{0.1601} & \underline{0.2313} & \underline{0.3239} \\
SNGP & 0.0750 & 0.0753 & 0.1717 & 0.1229 & 0.2012 & -0.0674 & -0.0552 & -0.0251 & 0.0507 & 0.1884 \\
\textbf{AdaKerNet} & \underline{0.2077} & \textbf{0.3105} & \textbf{0.3856} & \textbf{0.4349} & \textbf{0.4888} & \textbf{0.1206} & \textbf{0.1916} & \textbf{0.2812} & \textbf{0.3318} & \textbf{0.3957} \\
Avg.\ err.\ red. & \textcolor{gaingreen}{\textbf{$+$22.1\%}} & \textcolor{gaingreen}{\textbf{$+$24.5\%}} & \textcolor{gaingreen}{\textbf{$+$22.5\%}} & \textcolor{gaingreen}{\textbf{$+$23.0\%}} & \textcolor{gaingreen}{\textbf{$+$20.6\%}} & \textcolor{gaingreen}{\textbf{$+$26.5\%}} & \textcolor{gaingreen}{\textbf{$+$25.3\%}} & \textcolor{gaingreen}{\textbf{$+$23.6\%}} & \textcolor{gaingreen}{\textbf{$+$20.8\%}} & \textcolor{gaingreen}{\textbf{$+$17.5\%}} \\[2pt]
\midrule
\rule{0pt}{15pt} & & & \makebox[0pt]{\textbf{AMAZON-FASHION / Qwen}} & & & & & \makebox[0pt]{\textbf{QUECHUA-VALENCE / Gemini~2}} & & \\[4pt]
MLP & \underline{0.2377} & \underline{0.2630} & \underline{0.3179} & \underline{0.3155} & 0.3616 & \underline{-0.0545} & -0.1258 & -0.0790 & -0.0448 & 0.0361 \\
Transformer & 0.1871 & 0.2070 & 0.2483 & 0.2322 & 0.3448 & -0.1177 & -0.1864 & -0.0972 & -0.0479 & 0.0542 \\
AE $+$ linear & -2.8402 & -0.8798 & -0.1043 & 0.2027 & 0.3527 & -3.7288 & -1.1900 & -0.2295 & 0.0669 & 0.2018 \\
KRR & -0.0572 & 0.1170 & 0.2019 & 0.2898 & \underline{0.3859} & -0.0696 & \underline{0.0693} & \underline{0.1264} & \underline{0.1868} & \underline{0.2609} \\
SNGP & 0.1663 & 0.1691 & 0.1841 & 0.1322 & 0.2191 & -0.0594 & -0.1129 & -0.0676 & -0.0010 & 0.2258 \\
\textbf{AdaKerNet} & \textbf{0.2618} & \textbf{0.3337} & \textbf{0.3920} & \textbf{0.4261} & \textbf{0.4782} & \textbf{0.0620} & \textbf{0.1756} & \textbf{0.2108} & \textbf{0.2349} & \textbf{0.2719} \\
Avg.\ err.\ red. & \textcolor{gaingreen}{\textbf{$+$27.0\%}} & \textcolor{gaingreen}{\textbf{$+$26.9\%}} & \textcolor{gaingreen}{\textbf{$+$24.8\%}} & \textcolor{gaingreen}{\textbf{$+$24.5\%}} & \textcolor{gaingreen}{\textbf{$+$21.2\%}} & \textcolor{gaingreen}{\textbf{$+$26.2\%}} & \textcolor{gaingreen}{\textbf{$+$31.4\%}} & \textcolor{gaingreen}{\textbf{$+$25.3\%}} & \textcolor{gaingreen}{\textbf{$+$20.2\%}} & \textcolor{gaingreen}{\textbf{$+$12.7\%}} \\[2pt]
\midrule
\rule{0pt}{15pt} & & & \makebox[0pt]{\textbf{SPEECHOCEAN762-FLUENCY / Gemini~2}} & & & & & \makebox[0pt]{\textbf{SPEECHOCEAN762-PROSODIC / Gemini~2}} & & \\[4pt]
MLP & \textbf{0.2092} & \underline{0.2226} & \underline{0.2454} & 0.2593 & 0.3611 & \underline{0.2226} & \underline{0.2427} & 0.2622 & 0.2844 & 0.3740 \\
Transformer & 0.1973 & 0.1963 & 0.2077 & 0.2440 & 0.3250 & 0.2038 & 0.2196 & 0.2381 & 0.2458 & 0.3314 \\
AE $+$ linear & -1.5754 & -0.6767 & -0.0158 & 0.1911 & 0.3039 & -1.3843 & -0.5692 & 0.0351 & 0.2249 & 0.3289 \\
KRR & -0.2134 & 0.0791 & 0.1784 & \underline{0.2829} & \underline{0.3699} & 0.0065 & 0.2028 & \underline{0.2699} & \underline{0.3380} & \underline{0.4033} \\
SNGP & 0.1230 & 0.1356 & 0.1648 & 0.1513 & 0.2463 & 0.1658 & 0.1804 & 0.1879 & 0.1839 & 0.2628 \\
\textbf{AdaKerNet} & \underline{0.2033} & \textbf{0.3321} & \textbf{0.3568} & \textbf{0.3963} & \textbf{0.4204} & \textbf{0.2230} & \textbf{0.3437} & \textbf{0.3515} & \textbf{0.3873} & \textbf{0.4155} \\
Avg.\ err.\ red. & \textcolor{gaingreen}{\textbf{$+$22.5\%}} & \textcolor{gaingreen}{\textbf{$+$28.3\%}} & \textcolor{gaingreen}{\textbf{$+$23.0\%}} & \textcolor{gaingreen}{\textbf{$+$21.7\%}} & \textcolor{gaingreen}{\textbf{$+$14.3\%}} & \textcolor{gaingreen}{\textbf{$+$19.7\%}} & \textcolor{gaingreen}{\textbf{$+$25.0\%}} & \textcolor{gaingreen}{\textbf{$+$18.2\%}} & \textcolor{gaingreen}{\textbf{$+$17.3\%}} & \textcolor{gaingreen}{\textbf{$+$11.0\%}} \\
\bottomrule
\end{tabular}%
}
\end{table*}

\subsubsection{Main Results}

Table~\ref{tab:main} compares AdaKerNet against baselines B-I--B-V across six continuous prediction benchmarks, four pretrained MLLM encoders, and five label budgets, yielding 60 benchmark--MLLM--label-budget configurations. For PAD-UFES-20-MOLE-LESION, the largest label budget corresponds to the full labeled pool of $N=1043$ samples; it is reported under the $N=1000$ column for compactness. To demonstrate the versatility of AdaKerNet across a diverse set of reference kernels, we report results using an RBF kernel for PAD-UFES-20-AGE, PAD-UFES-20-MOLE-LESION, and SPEECHOCEAN762-FLUENCY; a Matérn kernel with $\nu=1/2$ for AMAZON FASHION; a Matérn kernel with $\nu=5/2$ for QUECHUA-VALENCE; and a Matérn kernel with $\nu=3/2$ for SPEECHOCEAN762-PROSODIC. The supplementary material further evaluates AdaKerNet under alternative reference-kernel choices at each benchmark.   AdaKerNet achieves the highest mean test $R^2$ in 58 configurations and ranks second in the remaining two. It outperforms the attention-based decoder (B-II), autoencoder with a linear head (B-III), fixed-kernel KRR (B-IV), and SNGP (B-V) in \textit{every} configuration, and the direct MLP decoder (B-I) in 58 of 60. Its relative reduction in test MSE, averaged over the five baselines, ranges from $7.8\%$ to $41.4\%$, with the largest reduction observed on PAD-UFES-20-AGE using Qwen2.5-VL at $N=100$. The gains over conventional neural decoders (B-I--B-II) are particularly pronounced when direct prediction from frozen representations is ineffective. On PAD-UFES-20-AGE with Qwen2.5-VL and $N=300$, AdaKerNet achieves $R^2=0.3522$, compared with $0.1214$ for B-I and $-0.0162$ for B-II. On QUECHUA-VALENCE, it obtains positive $R^2$ at every label budget, whereas both baselines yield negative values for $N\leq500$. Benefits also persist when direct prediction is effective: on SPEECHOCEAN762-FLUENCY with $N=200$, AdaKerNet achieves $R^2=0.3321$, compared to $0.2226$ and $0.1963$ for B-I and B-II, respectively. The consistent gains over B-III--B-V further demonstrate that neither generic autoencoder compression nor fixed-kernel KRR or SNGP applied directly to frozen representations matches AdaKerNet's performance in these experiments. 

Overall, the only exceptions occur at $N=100$ on AMAZON-FASHION with BLIP-2 and SPEECHOCEAN762-FLUENCY, where the direct MLP performs slightly better, by $0.0075$ and $0.0059$ in $R^2$, respectively.  Our intuition is that in these two settings, an MLP operating on the Lipschitz-controlled features also underperforms direct prediction from the frozen representations
(c.f. Table 3 of the supplementary material), while AdaKerNet partially closes this gap. This pattern suggests that, at the smallest label budget in these two benchmarks, the first Lipschitz-controlled feature extraction stage may not yet enhance the predictive structure already present in the frozen representations. Note though that this behavior is not representative of the broader results as AdaKerNet surpasses all five baselines at every larger budget on both benchmark--MLLM pairs. Overall, these results support the effectiveness of combining Lipschitz-controlled feature learning with task-adaptive neural kernel decoding across diverse multimodal prediction tasks under limited supervision.

It is also worth noticing that AdaKerNet continues to outperform B-IV (KRR Head) when alternative kernel choices are considered for the baseline. We show in Table 21 of the supplementary material that AdaKerNet still outperforms B-IV (now using a Mat\'ern-$1/2$ kernel instead of RBF) in all 60 reported configurations, across the evaluated MLLMs and label budgets. These
results indicate that its advantage over KRR extends beyond the RBF kernel choice. 

\vspace{-0.25cm}

\subsubsection{Ablation Studies}

We provide a broad set of additional ablation studies to further validate the merits of AdaKerNet and analyze its behavior. Due to space limitations, we highlight only the key findings here, while detailed results and discussion are provided in the supplementary material.

Initially, beyond evaluating the overall effectiveness of AdaKerNet against competing baselines, we seek to isolate and contextualize the contribution of each of its structural components. Specifically, we investigate (i) the role of the Lipschitz-controlled feature extraction block and (ii) the complementary benefits of the kernel reconstruction and supervised neural prediction losses within a \textit{unified} joint optimization framework.

\textit{Benefits of Lipschitz-controlled features.}
The significance of using the Lipschitz-controlled features $\vz$ instead of the frozen representations $\ve$ is demonstrated in Table 3 of the supplementary material, where the corresponding downstream predictor (MLP and KRR) architecture is retained, apart from its input dimensionality; the hidden layers and activations in MLP, and kernel form in KRR remain the same. Using $\vz$ improves mean test $R^2$ in 33 of 35 evaluated configurations for MLP and 34 of 35 for KRR, with relative test MSE reductions reaching $28.5\%$ and $38.7\%$, respectively. 
These results showcase the merits of the Lipschitz-controlled feature extractor block for both neural MLP and kernel-based downstream prediction.

\textit{Benefits of joint neural predictor and kernel representation learning.} Using $\vz$ alone does not fully account for AdaKerNet's performance. Table~4 in the supplementary material evaluates the contribution of AdaKerNet beyond the Lipschitz-controlled features $\vz$ by comparing the complete (end-to-end) decoder with a neural predictor (MLP) trained directly on the same features $\vz$. AdaKerNet achieves higher mean test $R^2$ in 54 of 60 configurations, with relative $R^2$ gains reaching $25.2\%$ and the shortfalls being at most $0.0040$. 

Table 5 in the supplementary material further compares AdaKerNet with an architecture-matched deeper MLP on $\vz$, obtained by removing the kernel-reconstruction loss ($\mathcal{L}_{\mathrm{ker}}=0$). AdaKerNet achieves higher mean test $R^2$ in 53 of 60 configurations, with reported relative gains reaching $30.2\%$ and shortfalls of at most $0.0059$, supporting the contribution of kernel-reconstruction regularization beyond network depth alone. These results support the additional benefit of jointly learning the kernel representation and neural predictor over direct MLP neural prediction from $\vz$. \\
To further assess the benefits of the deformed kernel $\widehat{k}_\psi$ (learned through joint training) over the reference kernel $k_0$, Tables~6--11 of the supplementary material compare KRR heads using these two kernels across different benchmarks, MLLMs, reference-kernel choices, and label budgets, while keeping the learned features $\vz$ fixed within each comparison. The learned kernel yields higher $R^2$ in 217 of the 240 configurations across four reference-kernel families, indicating that the benefits of the learned similarity function through adaptive kernel deformation extend beyond AdaKerNet's nonlinear prediction head. 

\textit{Comparison of AdaKerNet to the kernel learning TRF counterpart \cite{shilton2022trf}}.  To further assess the benefits of AdaKerNet's adaptive kernel
deformation stage, we compare it with Tuned Random Features
(TRF)~\citep{shilton2022trf}, which learns a translation-invariant kernel by optimizing spectral weights over sampled frequencies. Both methods operate on the same features $\vz$ and use the same reference kernel. Table 22 in the supplementary material shows that AdaKerNet achieves higher test $R^2$ in 26 of 30 configurations evaluated across 4 benchmarks and different MLLMs, with the largest gains at $N=100$, ranging from $17.4\%$ to $137.7\%$. Its four shortfalls are at most $0.0099$ in absolute $R^2$. These results support the benefits of jointly learning a flexible kernel feature map and a nonlinear neural predictor, particularly under scarce supervision.

In addition to the structural ablations above, we further investigate the following questions:

\textbf{Q1. \textit{How does AdaKerNet perform compared to Baselines B.I -- B.V if a different reference kernel is used}?} To assess sensitivity to the reference kernel, we evaluate AdaKerNet using RBF and Mat\'ern-$1/2$, $3/2$, and $5/2$
kernels on PAD-UFES-20-AGE prediction with three MLLM encoders and SPEECHOCEAN-FLUENCY prediction with Gemini Embedding 2. Across the resulting 80 configurations (with different reference kernel choices) shown in Tables 12-13, AdaKerNet outperforms the direct MLP baseline (B-I) in 78 configurations and the attention-based baseline (B-II) in all 80. AdaKerNet additionally outperforms B-III -- B-V in all settings for all different reference kernel types. 

\noindent\textbf{Q2: \textit{How does AdaKerNet perform in the non-sparse label regime}?} AdaKerNet remains effective beyond the sparse-label regime. Across the 16 configurations assessed in Tables 14-19 of the supplementary material, with label budgets ranging from 1,043 to 3,486, AdaKerNet outperforms all five baselines simultaneously in 12 configurations. These results show that AdaKerNet's benefits can also extend to larger label budgets.

\noindent\textbf{Q3: \textit{How does $\lambda$ in AdaKerNet affect performance}?} 
The parameter $\lambda$ balances reference-kernel reconstruction fidelity and task-specific adaptation: excessively small values may provide insufficient predictive supervision, whereas excessively large values weaken the relative influence of kernel reconstruction. As shown in Table~20 of the supplementary material for QUECHUA-VALENCE, $\lambda$ selected from $\{0.1,1,10\}$ yields the best result at every label budget and outperforms the extreme values in 10 of 12 kernel--budget configurations. This indicates that intermediate values of $\lambda$ generally outperform extremely low or high values, underscoring the importance of balancing kernel reconstruction and task-specific adaptation.

\section{Conclusions}

We introduced AdaKerNet, a novel neural kernel decoder for prediction from frozen MLLM representations when labeled data are limited. AdaKerNet (i) learns Lipschitz-controlled features whose geometry guides subsequent decoding steps and facilitates alignment with a reference kernel that can encode a suitable structural prior; and (ii) jointly learns a task-adaptive kernel representation and a neural prediction head. A reference kernel guides this learning, while prediction supervision adaptively deforms the learned similarities to the downstream task. The proposed decoder requires no access to the MLLM's parameters or fine-tuning of the underlying model. Numerical tests across six continuous prediction benchmarks and four MLLMs show that AdaKerNet outperforms all baselines in 58 of 60 evaluated configurations. Ablation studies support the benefits of both the Lipschitz-controlled features and the joint kernel representation and neural predictor learning. These results support AdaKerNet as an effective approach to multimodal prediction under limited supervision. Extending adaptive kernel deformation for decoders for additional downstream tasks and applications belongs to our future research agenda.

\newpage

\newpage

\section*{\centerline{\huge Supplementary File}}

\setcounter{table}{1}

\begin{table}[t]
    \centering
    \small
    \setlength{\tabcolsep}{4pt}
    \begin{tabular}{@{}llcccc@{}}
        \toprule
        \textbf{Dataset} & \textbf{Modalities} & \textbf{Target} &
        \textbf{Labeled pool} & \textbf{Test} &
        \textbf{$\sigma_y$}\\
        \midrule
        PAD-UFES-20-AGE & image, text, tabular & patient age &
        1{,}612 & 367 & 15.8\\
        PAD-UFES-20-MOLE-LESION & image, text, tabular & $\log$ lesion diameter &
        1{,}043 & 284 & 0.579\\
        AMAZON FASHION & image, text, tabular & $\log$ price &
        3{,}486 & 747 & 0.814\\
        QUECHUA-VALENCE  & audio, transcript & valence &
        2{,}800 & 600 & 0.207\\
        SPEECHOCEAN762-FLUENCY & audio, transcript & fluency &
        2{,}000 & 2{,}500 & 1.492\\
        SPEECHOCEAN762-PROSODIC & audio, transcript & prosodic &
        2{,}000 & 2{,}500 & 1.469\\
        \bottomrule
    \end{tabular}
    \caption{Multimodal regression benchmarks. $\sigma_y$ is the
    standard deviation of the target; $R^2$ is scale-free and therefore
    comparable across datasets, while absolute errors are not.}
    \label{tab:datasets}
\end{table}

\begin{table}[!t]
\centering
\definecolor{gaingreen}{RGB}{0,128,55}
\definecolor{lossred}{RGB}{190,30,45}
\caption{Effect of the Lipschitz-controlled features $\vz$ utilizing the test $R^2$ metric. For each MLP or KRR predictor, the first row uses the raw MLLM representations $\ve$ and the second the Lipschitz-controlled features $\vz$: \textsc{MLP} is the multilayer perceptron head of B-I, \textsc{KRR} on $\ve$ is B-IV, and \textsc{KRR with} $k_0$ applies kernel ridge regression to $\vz$ with the reference kernel $k_0$, which for these three benchmarks has the same kernel form (RBF) as used by B-IV. Each \emph{Err.\ red.}\ row gives the relative reduction (\%) in test MSE obtained by replacing $\ve$ with $\vz$, $(R^2_{\vz}-R^2_{\ve})/(1-R^2_{\ve})$.}
\label{tab:structural}
\footnotesize
\setlength{\aboverulesep}{0pt}\setlength{\belowrulesep}{0pt}
\begin{tabular*}{\linewidth}{@{\extracolsep{\fill}}lccccc@{}}
\toprule
\textbf{Method} & $N{=}100$ & $N{=}200$ & $N{=}300$ & $N{=}500$ & $N{=}1000$ \\
\midrule
\multicolumn{6}{c}{\rule{0pt}{11pt}\textbf{PAD-UFES-20-AGE / BLIP-2}} \\[1pt]
MLP on $\ve$ & 0.2586 & 0.2841 & 0.3359 & 0.3729 & 0.4129 \\
MLP on $\vz$ & 0.2755 & 0.3133 & 0.3494 & 0.3717 & 0.4168 \\
\quad Err.\ red. & \textcolor{gaingreen}{\textbf{$+$2.3\%}} & \textcolor{gaingreen}{\textbf{$+$4.1\%}} & \textcolor{gaingreen}{\textbf{$+$2.0\%}} & \textcolor{lossred}{$-$0.2\%} & \textcolor{gaingreen}{\textbf{$+$0.7\%}} \\
KRR on $\ve$ & 0.1338 & 0.2386 & 0.2898 & 0.3467 & 0.4120 \\
KRR with $k_0$ on $\vz$ & 0.2531 & 0.3036 & 0.3414 & 0.3776 & 0.4245 \\
\quad Err.\ red. & \textcolor{gaingreen}{\textbf{$+$13.8\%}} & \textcolor{gaingreen}{\textbf{$+$8.5\%}} & \textcolor{gaingreen}{\textbf{$+$7.3\%}} & \textcolor{gaingreen}{\textbf{$+$4.7\%}} & \textcolor{gaingreen}{\textbf{$+$2.1\%}} \\[2pt]
\midrule
\multicolumn{6}{c}{\rule{0pt}{11pt}\textbf{PAD-UFES-20-AGE / LLaVA}} \\[1pt]
MLP on $\ve$ & 0.2615 & 0.2925 & 0.3399 & 0.3511 & 0.4082 \\
MLP on $\vz$ & 0.2996 & 0.3484 & 0.3847 & 0.3897 & 0.4436 \\
\quad Err.\ red. & \textcolor{gaingreen}{\textbf{$+$5.2\%}} & \textcolor{gaingreen}{\textbf{$+$7.9\%}} & \textcolor{gaingreen}{\textbf{$+$6.8\%}} & \textcolor{gaingreen}{\textbf{$+$5.9\%}} & \textcolor{gaingreen}{\textbf{$+$6.0\%}} \\
KRR on $\ve$ & -0.1548 & 0.0680 & 0.1569 & 0.2577 & 0.3521 \\
KRR with $k_0$ on $\vz$ & 0.2924 & 0.3507 & 0.3879 & 0.3984 & 0.4459 \\
\quad Err.\ red. & \textcolor{gaingreen}{\textbf{$+$38.7\%}} & \textcolor{gaingreen}{\textbf{$+$30.3\%}} & \textcolor{gaingreen}{\textbf{$+$27.4\%}} & \textcolor{gaingreen}{\textbf{$+$19.0\%}} & \textcolor{gaingreen}{\textbf{$+$14.5\%}} \\[2pt]
\midrule
\multicolumn{6}{c}{\rule{0pt}{11pt}\textbf{PAD-UFES-20-AGE / Qwen}} \\[1pt]
MLP on $\ve$ & -0.0157 & 0.0691 & 0.1214 & 0.1150 & 0.1604 \\
MLP on $\vz$ & 0.2740 & 0.3196 & 0.3474 & 0.3545 & 0.3812 \\
\quad Err.\ red. & \textcolor{gaingreen}{\textbf{$+$28.5\%}} & \textcolor{gaingreen}{\textbf{$+$26.9\%}} & \textcolor{gaingreen}{\textbf{$+$25.7\%}} & \textcolor{gaingreen}{\textbf{$+$27.1\%}} & \textcolor{gaingreen}{\textbf{$+$26.3\%}} \\
KRR on $\ve$ & -0.1291 & 0.0347 & 0.1103 & 0.2012 & 0.2991 \\
KRR with $k_0$ on $\vz$ & 0.2611 & 0.3052 & 0.3398 & 0.3647 & 0.3941 \\
\quad Err.\ red. & \textcolor{gaingreen}{\textbf{$+$34.6\%}} & \textcolor{gaingreen}{\textbf{$+$28.0\%}} & \textcolor{gaingreen}{\textbf{$+$25.8\%}} & \textcolor{gaingreen}{\textbf{$+$20.5\%}} & \textcolor{gaingreen}{\textbf{$+$13.6\%}} \\[2pt]
\midrule
\multicolumn{6}{c}{\rule{0pt}{11pt}\textbf{PAD-UFES-20-MOLE-LESION / BLIP-2}} \\[1pt]
MLP on $\ve$ & -0.0173 & 0.0065 & 0.1067 & 0.1333 & 0.1856 \\
MLP on $\vz$ & 0.1618 & 0.2205 & 0.2636 & 0.3322 & 0.3630 \\
\quad Err.\ red. & \textcolor{gaingreen}{\textbf{$+$17.6\%}} & \textcolor{gaingreen}{\textbf{$+$21.5\%}} & \textcolor{gaingreen}{\textbf{$+$17.6\%}} & \textcolor{gaingreen}{\textbf{$+$22.9\%}} & \textcolor{gaingreen}{\textbf{$+$21.8\%}} \\
KRR on $\ve$ & -0.0072 & 0.1275 & 0.1882 & 0.2962 & 0.3659 \\
KRR with $k_0$ on $\vz$ & 0.1364 & 0.2110 & 0.2596 & 0.3240 & 0.3628 \\
\quad Err.\ red. & \textcolor{gaingreen}{\textbf{$+$14.3\%}} & \textcolor{gaingreen}{\textbf{$+$9.6\%}} & \textcolor{gaingreen}{\textbf{$+$8.8\%}} & \textcolor{gaingreen}{\textbf{$+$3.9\%}} & \textcolor{lossred}{$-$0.5\%} \\[2pt]
\midrule
\multicolumn{6}{c}{\rule{0pt}{11pt}\textbf{PAD-UFES-20-MOLE-LESION / LLaVA}} \\[1pt]
MLP on $\ve$ & -0.0518 & 0.0550 & 0.0978 & 0.1274 & 0.2134 \\
MLP on $\vz$ & 0.1403 & 0.2122 & 0.2471 & 0.2935 & 0.3286 \\
\quad Err.\ red. & \textcolor{gaingreen}{\textbf{$+$18.3\%}} & \textcolor{gaingreen}{\textbf{$+$16.6\%}} & \textcolor{gaingreen}{\textbf{$+$16.5\%}} & \textcolor{gaingreen}{\textbf{$+$19.0\%}} & \textcolor{gaingreen}{\textbf{$+$14.6\%}} \\
KRR on $\ve$ & -0.1783 & 0.0425 & 0.1157 & 0.2227 & 0.2971 \\
KRR with $k_0$ on $\vz$ & 0.1356 & 0.2096 & 0.2517 & 0.2954 & 0.3222 \\
\quad Err.\ red. & \textcolor{gaingreen}{\textbf{$+$26.6\%}} & \textcolor{gaingreen}{\textbf{$+$17.5\%}} & \textcolor{gaingreen}{\textbf{$+$15.4\%}} & \textcolor{gaingreen}{\textbf{$+$9.4\%}} & \textcolor{gaingreen}{\textbf{$+$3.6\%}} \\[2pt]
\midrule
\multicolumn{6}{c}{\rule{0pt}{11pt}\textbf{PAD-UFES-20-MOLE-LESION / Qwen}} \\[1pt]
MLP on $\ve$ & -0.0678 & -0.1302 & -0.0379 & -0.0288 & 0.0654 \\
MLP on $\vz$ & 0.1240 & 0.1658 & 0.1951 & 0.2399 & 0.2697 \\
\quad Err.\ red. & \textcolor{gaingreen}{\textbf{$+$18.0\%}} & \textcolor{gaingreen}{\textbf{$+$26.2\%}} & \textcolor{gaingreen}{\textbf{$+$22.4\%}} & \textcolor{gaingreen}{\textbf{$+$26.1\%}} & \textcolor{gaingreen}{\textbf{$+$21.9\%}} \\
KRR on $\ve$ & -0.0529 & 0.0301 & 0.0699 & 0.1479 & 0.2238 \\
KRR with $k_0$ on $\vz$ & 0.0796 & 0.1510 & 0.1954 & 0.2344 & 0.2703 \\
\quad Err.\ red. & \textcolor{gaingreen}{\textbf{$+$12.6\%}} & \textcolor{gaingreen}{\textbf{$+$12.5\%}} & \textcolor{gaingreen}{\textbf{$+$13.5\%}} & \textcolor{gaingreen}{\textbf{$+$10.2\%}} & \textcolor{gaingreen}{\textbf{$+$6.0\%}} \\[2pt]
\midrule
\multicolumn{6}{c}{\rule{0pt}{11pt}\textbf{SPEECHOCEAN762-FLUENCY / Gemini~2}} \\[1pt]
MLP on $\ve$ & 0.2092 & 0.2226 & 0.2454 & 0.2593 & 0.3611 \\
MLP on $\vz$ & 0.1957 & 0.3306 & 0.3529 & 0.3890 & 0.4142 \\
\quad Err.\ red. & \textcolor{lossred}{$-$1.7\%} & \textcolor{gaingreen}{\textbf{$+$13.9\%}} & \textcolor{gaingreen}{\textbf{$+$14.2\%}} & \textcolor{gaingreen}{\textbf{$+$17.5\%}} & \textcolor{gaingreen}{\textbf{$+$8.3\%}} \\
KRR on $\ve$ & -0.2134 & 0.0791 & 0.1784 & 0.2829 & 0.3699 \\
KRR with $k_0$ on $\vz$ & 0.1524 & 0.2840 & 0.3490 & 0.3457 & 0.3975 \\
\quad Err.\ red. & \textcolor{gaingreen}{\textbf{$+$30.1\%}} & \textcolor{gaingreen}{\textbf{$+$22.2\%}} & \textcolor{gaingreen}{\textbf{$+$20.8\%}} & \textcolor{gaingreen}{\textbf{$+$8.8\%}} & \textcolor{gaingreen}{\textbf{$+$4.4\%}} \\
\bottomrule
\end{tabular*}
\end{table}

\begin{table}[!t]
\centering
\definecolor{gaingreen}{RGB}{0,128,55}
\definecolor{lossred}{RGB}{190,30,45}
\caption{Significance of \textit{joint} task-adaptive neural kernel decoding on top of the Lipschitz-controlled features $\vz$ assessed using the test $R^2$ metric. \textsc{MLP on} $\vz$ applies a multilayer perceptron head to $\vz$; AdaKerNet is the full end-to-end model. The \emph{Rel.\ gain} row gives AdaKerNet's relative $R^2$ gain (\%) over MLP on $\vz$ as $(R^2_{\mathrm{AdaKerNet}}-R^2_{\mathrm{MLP}})/R^2_{\mathrm{MLP}}$.}
\label{tab:mlpz_vs_adakernet}
\setlength{\tabcolsep}{7pt}
\setlength{\aboverulesep}{0pt}\setlength{\belowrulesep}{0pt}
\resizebox{\linewidth}{!}{%
\begin{tabular}{@{}l*{5}{c}@{\hspace{6pt}}|@{\hspace{6pt}}*{5}{c}@{}}
\toprule
\textbf{Method} & $N{=}100$ & $200$ & $300$ & $500$ & $1000$ & $N{=}100$ & $200$ & $300$ & $500$ & $1000$ \\
\midrule
\rule{0pt}{15pt} & & & \makebox[0pt]{\textbf{PAD-UFES-20-AGE / BLIP-2}} & & & & & \makebox[0pt]{\textbf{PAD-UFES-20-AGE / LLaVA}} & & \\[4pt]
MLP on $\vz$ & 0.2755 & 0.3133 & 0.3494 & 0.3717 & 0.4168 & 0.2996 & 0.3484 & 0.3847 & 0.3897 & 0.4436 \\
AdaKerNet & 0.2750 & 0.3251 & 0.3579 & 0.3891 & 0.4261 & 0.3045 & 0.3588 & 0.3958 & 0.4061 & 0.4472 \\
\quad Rel.\ gain & \textcolor{lossred}{$-$0.2\%} & \textcolor{gaingreen}{\textbf{$+$3.8\%}} & \textcolor{gaingreen}{\textbf{$+$2.4\%}} & \textcolor{gaingreen}{\textbf{$+$4.7\%}} & \textcolor{gaingreen}{\textbf{$+$2.2\%}} & \textcolor{gaingreen}{\textbf{$+$1.6\%}} & \textcolor{gaingreen}{\textbf{$+$3.0\%}} & \textcolor{gaingreen}{\textbf{$+$2.9\%}} & \textcolor{gaingreen}{\textbf{$+$4.2\%}} & \textcolor{gaingreen}{\textbf{$+$0.8\%}} \\[2pt]
\midrule
\rule{0pt}{15pt} & & & \makebox[0pt]{\textbf{PAD-UFES-20-AGE / Qwen}} & & & & & \makebox[0pt]{\textbf{PAD-UFES-20-MOLE-LESION / BLIP-2}} & & \\[4pt]
MLP on $\vz$ & 0.2740 & 0.3196 & 0.3474 & 0.3545 & 0.3812 & 0.1618 & 0.2205 & 0.2636 & 0.3322 & 0.3630 \\
AdaKerNet & 0.2771 & 0.3167 & 0.3522 & 0.3690 & 0.3923 & 0.1777 & 0.2291 & 0.2731 & 0.3347 & 0.3716 \\
\quad Rel.\ gain & \textcolor{gaingreen}{\textbf{$+$1.1\%}} & \textcolor{lossred}{$-$0.9\%} & \textcolor{gaingreen}{\textbf{$+$1.4\%}} & \textcolor{gaingreen}{\textbf{$+$4.1\%}} & \textcolor{gaingreen}{\textbf{$+$2.9\%}} & \textcolor{gaingreen}{\textbf{$+$9.8\%}} & \textcolor{gaingreen}{\textbf{$+$3.9\%}} & \textcolor{gaingreen}{\textbf{$+$3.6\%}} & \textcolor{gaingreen}{\textbf{$+$0.8\%}} & \textcolor{gaingreen}{\textbf{$+$2.4\%}} \\[2pt]
\midrule
\rule{0pt}{15pt} & & & \makebox[0pt]{\textbf{PAD-UFES-20-MOLE-LESION / LLaVA}} & & & & & \makebox[0pt]{\textbf{PAD-UFES-20-MOLE-LESION / Qwen}} & & \\[4pt]
MLP on $\vz$ & 0.1403 & 0.2122 & 0.2471 & 0.2935 & 0.3286 & 0.1240 & 0.1658 & 0.1951 & 0.2399 & 0.2697 \\
AdaKerNet & 0.1475 & 0.2134 & 0.2497 & 0.2925 & 0.3246 & 0.1356 & 0.1731 & 0.1987 & 0.2380 & 0.2724 \\
\quad Rel.\ gain & \textcolor{gaingreen}{\textbf{$+$5.1\%}} & \textcolor{gaingreen}{\textbf{$+$0.6\%}} & \textcolor{gaingreen}{\textbf{$+$1.1\%}} & \textcolor{lossred}{$-$0.3\%} & \textcolor{lossred}{$-$1.2\%} & \textcolor{gaingreen}{\textbf{$+$9.4\%}} & \textcolor{gaingreen}{\textbf{$+$4.4\%}} & \textcolor{gaingreen}{\textbf{$+$1.8\%}} & \textcolor{lossred}{$-$0.8\%} & \textcolor{gaingreen}{\textbf{$+$1.0\%}} \\[2pt]
\midrule
\rule{0pt}{15pt} & & & \makebox[0pt]{\textbf{AMAZON-FASHION / BLIP-2}} & & & & & \makebox[0pt]{\textbf{AMAZON-FASHION / LLaVA}} & & \\[4pt]
MLP on $\vz$ & 0.1810 & 0.3022 & 0.3847 & 0.4337 & 0.4881 & 0.0963 & 0.1752 & 0.2804 & 0.3309 & 0.3922 \\
AdaKerNet & 0.2077 & 0.3105 & 0.3856 & 0.4349 & 0.4888 & 0.1206 & 0.1916 & 0.2812 & 0.3318 & 0.3957 \\
\quad Rel.\ gain & \textcolor{gaingreen}{\textbf{$+$14.8\%}} & \textcolor{gaingreen}{\textbf{$+$2.7\%}} & \textcolor{gaingreen}{\textbf{$+$0.2\%}} & \textcolor{gaingreen}{\textbf{$+$0.3\%}} & \textcolor{gaingreen}{\textbf{$+$0.1\%}} & \textcolor{gaingreen}{\textbf{$+$25.2\%}} & \textcolor{gaingreen}{\textbf{$+$9.4\%}} & \textcolor{gaingreen}{\textbf{$+$0.3\%}} & \textcolor{gaingreen}{\textbf{$+$0.3\%}} & \textcolor{gaingreen}{\textbf{$+$0.9\%}} \\[2pt]
\midrule
\rule{0pt}{15pt} & & & \makebox[0pt]{\textbf{AMAZON-FASHION / Qwen}} & & & & & \makebox[0pt]{\textbf{QUECHUA-VALENCE / Gemini~2}} & & \\[4pt]
MLP on $\vz$ & 0.2483 & 0.3291 & 0.3901 & 0.4227 & 0.4746 & 0.0511 & 0.1738 & 0.2103 & 0.2316 & 0.2657 \\
AdaKerNet & 0.2618 & 0.3337 & 0.3920 & 0.4261 & 0.4782 & 0.0620 & 0.1756 & 0.2108 & 0.2349 & 0.2719 \\
\quad Rel.\ gain & \textcolor{gaingreen}{\textbf{$+$5.4\%}} & \textcolor{gaingreen}{\textbf{$+$1.4\%}} & \textcolor{gaingreen}{\textbf{$+$0.5\%}} & \textcolor{gaingreen}{\textbf{$+$0.8\%}} & \textcolor{gaingreen}{\textbf{$+$0.8\%}} & \textcolor{gaingreen}{\textbf{$+$21.3\%}} & \textcolor{gaingreen}{\textbf{$+$1.0\%}} & \textcolor{gaingreen}{\textbf{$+$0.2\%}} & \textcolor{gaingreen}{\textbf{$+$1.4\%}} & \textcolor{gaingreen}{\textbf{$+$2.3\%}} \\[2pt]
\midrule
\rule{0pt}{15pt} & & & \makebox[0pt]{\textbf{SPEECHOCEAN762-FLUENCY / Gemini~2}} & & & & & \makebox[0pt]{\textbf{SPEECHOCEAN762-PROSODIC / Gemini~2}} & & \\[4pt]
MLP on $\vz$ & 0.1957 & 0.3306 & 0.3529 & 0.3890 & 0.4142 & 0.2096 & 0.3470 & 0.3491 & 0.3819 & 0.4090 \\
AdaKerNet & 0.2033 & 0.3321 & 0.3568 & 0.3963 & 0.4204 & 0.2230 & 0.3437 & 0.3515 & 0.3873 & 0.4155 \\
\quad Rel.\ gain & \textcolor{gaingreen}{\textbf{$+$3.9\%}} & \textcolor{gaingreen}{\textbf{$+$0.5\%}} & \textcolor{gaingreen}{\textbf{$+$1.1\%}} & \textcolor{gaingreen}{\textbf{$+$1.9\%}} & \textcolor{gaingreen}{\textbf{$+$1.5\%}} & \textcolor{gaingreen}{\textbf{$+$6.4\%}} & \textcolor{lossred}{$-$1.0\%} & \textcolor{gaingreen}{\textbf{$+$0.7\%}} & \textcolor{gaingreen}{\textbf{$+$1.4\%}} & \textcolor{gaingreen}{\textbf{$+$1.6\%}} \\
\bottomrule
\end{tabular}%
}
\end{table}

\begin{table}[!t]
\centering
\definecolor{gaingreen}{RGB}{0,128,55}
\definecolor{lossred}{RGB}{190,30,45}
\caption{Significance of \textit{joint} task-adaptive neural kernel decoding on top of the Lipschitz-controlled features $\vz$ assessed using the test $R^2$ metric. \textsc{MLP} on $\vz$ now applies a (deeper) multilayer perceptron head (corresponding to AdaKerNet with $\mathcal{L}_{\mathrm{ker}} = 0$) to $\vz$; AdaKerNet is the full end-to-end model. The \emph{Rel.\ gain} row gives AdaKerNet's relative $R^2$ gain (\%) over MLP on $\vz$ as $(R^2_{\mathrm{AdaKerNet}}-R^2_{\mathrm{MLP}})/R^2_{\mathrm{MLP}}$.}
\label{tab:deepmlpz_vs_adakernet}
\setlength{\tabcolsep}{7pt}
\setlength{\aboverulesep}{0pt}\setlength{\belowrulesep}{0pt}
\resizebox{\linewidth}{!}{%
\begin{tabular}{@{}l*{5}{c}@{\hspace{6pt}}|@{\hspace{6pt}}*{5}{c}@{}}
\toprule
\textbf{Method} & $N{=}100$ & $200$ & $300$ & $500$ & $1000$ & $N{=}100$ & $200$ & $300$ & $500$ & $1000$ \\
\midrule
\rule{0pt}{15pt} & & & \makebox[0pt]{\textbf{PAD-UFES-20-AGE / BLIP-2}} & & & & & \makebox[0pt]{\textbf{PAD-UFES-20-AGE / LLaVA}} & & \\[4pt]
Deep MLP (AdaKerNet - $\mathcal{L}_{\mathrm{ker}} = 0$) on $\vz$ & 0.2714 & 0.3129 & 0.3463 & 0.3720 & 0.4158 & 0.2998 & 0.3492 & 0.3869 & 0.3903 & 0.4434 \\
AdaKerNet & 0.2750 & 0.3251 & 0.3579 & 0.3891 & 0.4261 & 0.3045 & 0.3588 & 0.3958 & 0.4061 & 0.4472 \\
\quad Rel.\ gain & \textcolor{gaingreen}{\textbf{$+$1.3\%}} & \textcolor{gaingreen}{\textbf{$+$3.9\%}} & \textcolor{gaingreen}{\textbf{$+$3.3\%}} & \textcolor{gaingreen}{\textbf{$+$4.6\%}} & \textcolor{gaingreen}{\textbf{$+$2.5\%}} & \textcolor{gaingreen}{\textbf{$+$1.5\%}} & \textcolor{gaingreen}{\textbf{$+$2.8\%}} & \textcolor{gaingreen}{\textbf{$+$2.3\%}} & \textcolor{gaingreen}{\textbf{$+$4.0\%}} & \textcolor{gaingreen}{\textbf{$+$0.9\%}} \\[2pt]
\midrule
\rule{0pt}{15pt} & & & \makebox[0pt]{\textbf{PAD-UFES-20-AGE / Qwen}} & & & & & \makebox[0pt]{\textbf{PAD-UFES-20-MOLE-LESION / BLIP-2}} & & \\[4pt]
Deep MLP (AdaKerNet - $\mathcal{L}_{\mathrm{ker}} = 0$) on $\vz$ & 0.2700 & 0.3226 & 0.3483 & 0.3546 & 0.3820 & 0.1598 & 0.2215 & 0.2661 & 0.3344 & 0.3653 \\
AdaKerNet & 0.2771 & 0.3167 & 0.3522 & 0.3690 & 0.3923 & 0.1777 & 0.2291 & 0.2731 & 0.3347 & 0.3716 \\
\quad Rel.\ gain & \textcolor{gaingreen}{\textbf{$+$2.6\%}} & \textcolor{lossred}{$-$1.8\%} & \textcolor{gaingreen}{\textbf{$+$1.1\%}} & \textcolor{gaingreen}{\textbf{$+$4.1\%}} & \textcolor{gaingreen}{\textbf{$+$2.7\%}} & \textcolor{gaingreen}{\textbf{$+$11.1\%}} & \textcolor{gaingreen}{\textbf{$+$3.4\%}} & \textcolor{gaingreen}{\textbf{$+$2.6\%}} & \textcolor{gaingreen}{\textbf{$+$0.1\%}} & \textcolor{gaingreen}{\textbf{$+$1.7\%}} \\[2pt]
\midrule
\rule{0pt}{15pt} & & & \makebox[0pt]{\textbf{PAD-UFES-20-MOLE-LESION / LLaVA}} & & & & & \makebox[0pt]{\textbf{PAD-UFES-20-MOLE-LESION / Qwen}} & & \\[4pt]
Deep MLP (AdaKerNet - $\mathcal{L}_{\mathrm{ker}} = 0$) on $\vz$ & 0.1392 & 0.2146 & 0.2483 & 0.2928 & 0.3286 & 0.1197 & 0.1665 & 0.1946 & 0.2408 & 0.2714 \\
AdaKerNet & 0.1475 & 0.2134 & 0.2497 & 0.2925 & 0.3246 & 0.1356 & 0.1731 & 0.1987 & 0.2380 & 0.2724 \\
\quad Rel.\ gain & \textcolor{gaingreen}{\textbf{$+$6.0\%}} & \textcolor{lossred}{$-$0.5\%} & \textcolor{gaingreen}{\textbf{$+$0.6\%}} & \textcolor{lossred}{$-$0.1\%} & \textcolor{lossred}{$-$1.2\%} & \textcolor{gaingreen}{\textbf{$+$13.3\%}} & \textcolor{gaingreen}{\textbf{$+$3.9\%}} & \textcolor{gaingreen}{\textbf{$+$2.1\%}} & \textcolor{lossred}{$-$1.2\%} & \textcolor{gaingreen}{\textbf{$+$0.3\%}} \\[2pt]
\midrule
\rule{0pt}{15pt} & & & \makebox[0pt]{\textbf{AMAZON-FASHION / BLIP-2}} & & & & & \makebox[0pt]{\textbf{AMAZON-FASHION / LLaVA}} & & \\[4pt]
Deep MLP (AdaKerNet - $\mathcal{L}_{\mathrm{ker}} = 0$) on $\vz$ & 0.1779 & 0.3003 & 0.3852 & 0.4336 & 0.4885 & 0.0933 & 0.1715 & 0.2810 & 0.3317 & 0.3932 \\
AdaKerNet & 0.2077 & 0.3105 & 0.3856 & 0.4349 & 0.4888 & 0.1206 & 0.1916 & 0.2812 & 0.3318 & 0.3957 \\
\quad Rel.\ gain & \textcolor{gaingreen}{\textbf{$+$16.8\%}} & \textcolor{gaingreen}{\textbf{$+$3.4\%}} & \textcolor{gaingreen}{\textbf{$+$0.1\%}} & \textcolor{gaingreen}{\textbf{$+$0.3\%}} & \textcolor{gaingreen}{\textbf{$+$0.1\%}} & \textcolor{gaingreen}{\textbf{$+$29.3\%}} & \textcolor{gaingreen}{\textbf{$+$11.7\%}} & \textcolor{gaingreen}{\textbf{$+$0.1\%}} & \textcolor{gaingreen}{\textbf{$+$0.0\%}} & \textcolor{gaingreen}{\textbf{$+$0.6\%}} \\[2pt]
\midrule
\rule{0pt}{15pt} & & & \makebox[0pt]{\textbf{AMAZON-FASHION / Qwen}} & & & & & \makebox[0pt]{\textbf{QUECHUA-VALENCE / Gemini~2}} & & \\[4pt]
Deep MLP (AdaKerNet - $\mathcal{L}_{\mathrm{ker}} = 0$) on $\vz$ & 0.2436 & 0.3288 & 0.3909 & 0.4243 & 0.4752 & 0.0476 & 0.1744 & 0.2100 & 0.2304 & 0.2651 \\
AdaKerNet & 0.2618 & 0.3337 & 0.3920 & 0.4261 & 0.4782 & 0.0620 & 0.1756 & 0.2108 & 0.2349 & 0.2719 \\
\quad Rel.\ gain & \textcolor{gaingreen}{\textbf{$+$7.5\%}} & \textcolor{gaingreen}{\textbf{$+$1.5\%}} & \textcolor{gaingreen}{\textbf{$+$0.3\%}} & \textcolor{gaingreen}{\textbf{$+$0.4\%}} & \textcolor{gaingreen}{\textbf{$+$0.6\%}} & \textcolor{gaingreen}{\textbf{$+$30.2\%}} & \textcolor{gaingreen}{\textbf{$+$0.7\%}} & \textcolor{gaingreen}{\textbf{$+$0.4\%}} & \textcolor{gaingreen}{\textbf{$+$2.0\%}} & \textcolor{gaingreen}{\textbf{$+$2.6\%}} \\[2pt]
\midrule
\rule{0pt}{15pt} & & & \makebox[0pt]{\textbf{SPEECHOCEAN762-FLUENCY / Gemini~2}} & & & & & \makebox[0pt]{\textbf{SPEECHOCEAN762-PROSODIC / Gemini~2}} & & \\[4pt]
Deep MLP (AdaKerNet - $\mathcal{L}_{\mathrm{ker}} = 0$) on $\vz$ & 0.1921 & 0.3275 & 0.3524 & 0.3898 & 0.4137 & 0.2023 & 0.3478 & 0.3524 & 0.3826 & 0.4091 \\
AdaKerNet & 0.2033 & 0.3321 & 0.3568 & 0.3963 & 0.4204 & 0.2230 & 0.3437 & 0.3515 & 0.3873 & 0.4155 \\
\quad Rel.\ gain & \textcolor{gaingreen}{\textbf{$+$5.8\%}} & \textcolor{gaingreen}{\textbf{$+$1.4\%}} & \textcolor{gaingreen}{\textbf{$+$1.3\%}} & \textcolor{gaingreen}{\textbf{$+$1.7\%}} & \textcolor{gaingreen}{\textbf{$+$1.6\%}} & \textcolor{gaingreen}{\textbf{$+$10.2\%}} & \textcolor{lossred}{$-$1.2\%} & \textcolor{lossred}{$-$0.2\%} & \textcolor{gaingreen}{\textbf{$+$1.2\%}} & \textcolor{gaingreen}{\textbf{$+$1.6\%}} \\
\bottomrule
\end{tabular}%
}
\end{table}

\begin{table}[t]
\centering
\definecolor{gaingreen}{RGB}{0,128,55}
\definecolor{lossred}{RGB}{190,30,45}
\caption{PAD-UFES-20-AGE benchmark. Kernel ridge regression (KRR) with the learned kernel $\widehat{k}_\psi$ against KRR with the fixed reference kernel $k_0$ based on the test $R^2$ metric. The two share the same Lipschitz-controlled feature vector $\vz$ and the same reference kernel, and both select the ridge parameter based on the $N$ labeled points. The \emph{Rel.\ gain} (\%) row is the relative $R^2$ gain of KRR with $\hat{k}_{\psi}$ over KRR with $k_0$, as $\,(R^2_{\hat{k}_{\psi}}-R^2_{k_0})/R^2_{\mathrm{k_0}}$.}
\label{tab:q1padufes20}
\footnotesize
\setlength{\tabcolsep}{4pt}
\begin{tabular}{@{}lllccccc@{}}
\toprule
\textbf{Encoder} & \textbf{Kernel} & \textbf{Predictor} & $N=100$ & $N=200$ & $N=300$ & $N=500$ & $N=1000$ \\
\midrule
\multirow{3}{*}{BLIP-2} & \multirow{3}{*}{Mat\'ern-$1/2$} & $\widehat{k}_\psi$  & 0.2571 & 0.3132 & 0.3573 & 0.3917 & 0.4244 \\
 &  & $k_0$  & 0.2545 & 0.3025 & 0.3452 & 0.3779 & 0.4255 \\
 &  & Rel.\ gain & \textcolor{gaingreen}{\textbf{+1.01\%}} & \textcolor{gaingreen}{\textbf{+3.53\%}} & \textcolor{gaingreen}{\textbf{+3.49\%}} & \textcolor{gaingreen}{\textbf{+3.66\%}} & \textcolor{lossred}{-0.27\%} \\
\cmidrule{2-8}
 & \multirow{3}{*}{Mat\'ern-$3/2$} & $\widehat{k}_\psi$  & 0.2592 & 0.3152 & 0.3571 & 0.3878 & 0.4279 \\
 &  & $k_0$  & 0.2524 & 0.3041 & 0.3449 & 0.3787 & 0.4256 \\
 &  & Rel.\ gain & \textcolor{gaingreen}{\textbf{+2.68\%}} & \textcolor{gaingreen}{\textbf{+3.63\%}} & \textcolor{gaingreen}{\textbf{+3.54\%}} & \textcolor{gaingreen}{\textbf{+2.42\%}} & \textcolor{gaingreen}{\textbf{+0.55\%}} \\
\cmidrule{2-8}
 & \multirow{3}{*}{Mat\'ern-$5/2$} & $\widehat{k}_\psi$  & 0.2618 & 0.3187 & 0.3549 & 0.3859 & 0.4263 \\
 &  & $k_0$  & 0.2551 & 0.3056 & 0.3448 & 0.3793 & 0.4261 \\
 &  & Rel.\ gain & \textcolor{gaingreen}{\textbf{+2.66\%}} & \textcolor{gaingreen}{\textbf{+4.28\%}} & \textcolor{gaingreen}{\textbf{+2.94\%}} & \textcolor{gaingreen}{\textbf{+1.73\%}} & \textcolor{gaingreen}{\textbf{+0.04\%}} \\
\cmidrule{2-8}
 & \multirow{3}{*}{RBF} & $\widehat{k}_\psi$  & 0.2661 & 0.3176 & 0.3546 & 0.3857 & 0.4249 \\
 &  & $k_0$  & 0.2531 & 0.3036 & 0.3414 & 0.3776 & 0.4245 \\
 &  & Rel.\ gain & \textcolor{gaingreen}{\textbf{+5.15\%}} & \textcolor{gaingreen}{\textbf{+4.62\%}} & \textcolor{gaingreen}{\textbf{+3.86\%}} & \textcolor{gaingreen}{\textbf{+2.12\%}} & \textcolor{gaingreen}{\textbf{+0.10\%}} \\
\midrule
\multirow{3}{*}{LLaVA} & \multirow{3}{*}{Mat\'ern-$1/2$} & $\widehat{k}_\psi$  & 0.2961 & 0.3613 & 0.3967 & 0.4047 & 0.4458 \\
 &  & $k_0$  & 0.2974 & 0.3517 & 0.3943 & 0.3997 & 0.4444 \\
 &  & Rel.\ gain & \textcolor{lossred}{-0.43\%} & \textcolor{gaingreen}{\textbf{+2.71\%}} & \textcolor{gaingreen}{\textbf{+0.61\%}} & \textcolor{gaingreen}{\textbf{+1.26\%}} & \textcolor{gaingreen}{\textbf{+0.31\%}} \\
\cmidrule{2-8}
 & \multirow{3}{*}{Mat\'ern-$3/2$} & $\widehat{k}_\psi$  & 0.2946 & 0.3576 & 0.3969 & 0.4042 & 0.4450 \\
 &  & $k_0$  & 0.2915 & 0.3501 & 0.3917 & 0.3990 & 0.4455 \\
 &  & Rel.\ gain & \textcolor{gaingreen}{\textbf{+1.07\%}} & \textcolor{gaingreen}{\textbf{+2.14\%}} & \textcolor{gaingreen}{\textbf{+1.31\%}} & \textcolor{gaingreen}{\textbf{+1.30\%}} & \textcolor{lossred}{-0.12\%} \\
\cmidrule{2-8}
 & \multirow{3}{*}{Mat\'ern-$5/2$} & $\widehat{k}_\psi$  & 0.2944 & 0.3543 & 0.3979 & 0.4022 & 0.4469 \\
 &  & $k_0$  & 0.2930 & 0.3503 & 0.3902 & 0.3990 & 0.4462 \\
 &  & Rel.\ gain & \textcolor{gaingreen}{\textbf{+0.47\%}} & \textcolor{gaingreen}{\textbf{+1.15\%}} & \textcolor{gaingreen}{\textbf{+1.97\%}} & \textcolor{gaingreen}{\textbf{+0.79\%}} & \textcolor{gaingreen}{\textbf{+0.15\%}} \\
\cmidrule{2-8}
 & \multirow{3}{*}{RBF} & $\widehat{k}_\psi$  & 0.3006 & 0.3587 & 0.3977 & 0.4024 & 0.4462 \\
 &  & $k_0$  & 0.2924 & 0.3507 & 0.3879 & 0.3984 & 0.4459 \\
 &  & Rel.\ gain & \textcolor{gaingreen}{\textbf{+2.82\%}} & \textcolor{gaingreen}{\textbf{+2.28\%}} & \textcolor{gaingreen}{\textbf{+2.51\%}} & \textcolor{gaingreen}{\textbf{+1.02\%}} & \textcolor{gaingreen}{\textbf{+0.07\%}} \\
\midrule
\multirow{3}{*}{Qwen} & \multirow{3}{*}{Mat\'ern-$1/2$} & $\widehat{k}_\psi$  & 0.2701 & 0.3151 & 0.3528 & 0.3636 & 0.3894 \\
 &  & $k_0$  & 0.2628 & 0.3007 & 0.3445 & 0.3605 & 0.3966 \\
 &  & Rel.\ gain & \textcolor{gaingreen}{\textbf{+2.78\%}} & \textcolor{gaingreen}{\textbf{+4.80\%}} & \textcolor{gaingreen}{\textbf{+2.40\%}} & \textcolor{gaingreen}{\textbf{+0.86\%}} & \textcolor{lossred}{-1.80\%} \\
\cmidrule{2-8}
 & \multirow{3}{*}{Mat\'ern-$3/2$} & $\widehat{k}_\psi$  & 0.2730 & 0.3168 & 0.3531 & 0.3687 & 0.3909 \\
 &  & $k_0$  & 0.2597 & 0.3009 & 0.3442 & 0.3622 & 0.3954 \\
 &  & Rel.\ gain & \textcolor{gaingreen}{\textbf{+5.15\%}} & \textcolor{gaingreen}{\textbf{+5.28\%}} & \textcolor{gaingreen}{\textbf{+2.58\%}} & \textcolor{gaingreen}{\textbf{+1.80\%}} & \textcolor{lossred}{-1.16\%} \\
\cmidrule{2-8}
 & \multirow{3}{*}{Mat\'ern-$5/2$} & $\widehat{k}_\psi$  & 0.2668 & 0.3138 & 0.3533 & 0.3671 & 0.3900 \\
 &  & $k_0$  & 0.2608 & 0.3031 & 0.3428 & 0.3634 & 0.3949 \\
 &  & Rel.\ gain & \textcolor{gaingreen}{\textbf{+2.29\%}} & \textcolor{gaingreen}{\textbf{+3.54\%}} & \textcolor{gaingreen}{\textbf{+3.07\%}} & \textcolor{gaingreen}{\textbf{+1.03\%}} & \textcolor{lossred}{-1.24\%} \\
\cmidrule{2-8}
 & \multirow{3}{*}{RBF} & $\widehat{k}_\psi$  & 0.2642 & 0.3144 & 0.3529 & 0.3683 & 0.3924 \\
 &  & $k_0$  & 0.2611 & 0.3052 & 0.3398 & 0.3647 & 0.3941 \\
 &  & Rel.\ gain & \textcolor{gaingreen}{\textbf{+1.19\%}} & \textcolor{gaingreen}{\textbf{+3.01\%}} & \textcolor{gaingreen}{\textbf{+3.85\%}} & \textcolor{gaingreen}{\textbf{+1.01\%}} & \textcolor{lossred}{-0.43\%} \\
\bottomrule
\end{tabular}
\end{table}

\begin{table}[t]
\centering
\definecolor{gaingreen}{RGB}{0,128,55}
\definecolor{lossred}{RGB}{190,30,45}
\caption{PAD-UFES-20-MOLE-LESION benchmark. Kernel ridge regression (KRR) with the learned kernel $\widehat{k}_\psi$ against KRR with the fixed reference kernel $k_0$ based on the test $R^2$ metric. The two share the same Lipschitz-controlled feature vector $\vz$ and the same reference kernel, and both select the ridge parameter based on the $N$ labeled points. The \emph{Rel.\ gain} (\%) row is the relative $R^2$ gain of KRR with $\hat{k}_{\psi}$ over KRR with $k_0$, as $\,(R^2_{\hat{k}_{\psi}}-R^2_{k_0})/R^2_{\mathrm{k_0}}$.}
\label{tab:q1padufes20diameter}
\footnotesize
\setlength{\tabcolsep}{4pt}
\begin{tabular}{@{}lllccccc@{}}
\toprule
\textbf{Encoder} & \textbf{Kernel} & \textbf{Predictor} & $N=100$ & $N=200$ & $N=300$ & $N=500$ & $N=1043$ \\
\midrule
\multirow{3}{*}{BLIP-2} & \multirow{3}{*}{Mat\'ern-$1/2$} & $\widehat{k}_\psi$  & 0.1818 & 0.2241 & 0.2718 & 0.3354 & 0.3723 \\
 &  & $k_0$  & 0.1447 & 0.2167 & 0.2614 & 0.3303 & 0.3765 \\
 &  & Rel.\ gain & \textcolor{gaingreen}{\textbf{+25.57\%}} & \textcolor{gaingreen}{\textbf{+3.43\%}} & \textcolor{gaingreen}{\textbf{+3.97\%}} & \textcolor{gaingreen}{\textbf{+1.55\%}} & \textcolor{lossred}{-1.13\%} \\
\cmidrule{2-8}
 & \multirow{3}{*}{Mat\'ern-$3/2$} & $\widehat{k}_\psi$  & 0.1830 & 0.2280 & 0.2708 & 0.3356 & 0.3719 \\
 &  & $k_0$  & 0.1433 & 0.2151 & 0.2604 & 0.3291 & 0.3713 \\
 &  & Rel.\ gain & \textcolor{gaingreen}{\textbf{+27.71\%}} & \textcolor{gaingreen}{\textbf{+6.02\%}} & \textcolor{gaingreen}{\textbf{+4.03\%}} & \textcolor{gaingreen}{\textbf{+1.96\%}} & \textcolor{gaingreen}{\textbf{+0.14\%}} \\
\cmidrule{2-8}
 & \multirow{3}{*}{Mat\'ern-$5/2$} & $\widehat{k}_\psi$  & 0.1744 & 0.2311 & 0.2744 & 0.3374 & 0.3744 \\
 &  & $k_0$  & 0.1444 & 0.2149 & 0.2610 & 0.3281 & 0.3689 \\
 &  & Rel.\ gain & \textcolor{gaingreen}{\textbf{+20.83\%}} & \textcolor{gaingreen}{\textbf{+7.55\%}} & \textcolor{gaingreen}{\textbf{+5.15\%}} & \textcolor{gaingreen}{\textbf{+2.84\%}} & \textcolor{gaingreen}{\textbf{+1.49\%}} \\
\cmidrule{2-8}
 & \multirow{3}{*}{RBF} & $\widehat{k}_\psi$  & 0.1758 & 0.2307 & 0.2717 & 0.3347 & 0.3732 \\
 &  & $k_0$  & 0.1364 & 0.2110 & 0.2596 & 0.3240 & 0.3628 \\
 &  & Rel.\ gain & \textcolor{gaingreen}{\textbf{+28.89\%}} & \textcolor{gaingreen}{\textbf{+9.35\%}} & \textcolor{gaingreen}{\textbf{+4.65\%}} & \textcolor{gaingreen}{\textbf{+3.29\%}} & \textcolor{gaingreen}{\textbf{+2.87\%}} \\
\midrule
\multirow{3}{*}{LLaVA} & \multirow{3}{*}{Mat\'ern-$1/2$} & $\widehat{k}_\psi$  & 0.1539 & 0.2137 & 0.2505 & 0.2932 & 0.3249 \\
 &  & $k_0$  & 0.1295 & 0.2030 & 0.2481 & 0.2924 & 0.3243 \\
 &  & Rel.\ gain & \textcolor{gaingreen}{\textbf{+18.83\%}} & \textcolor{gaingreen}{\textbf{+5.29\%}} & \textcolor{gaingreen}{\textbf{+0.97\%}} & \textcolor{gaingreen}{\textbf{+0.26\%}} & \textcolor{gaingreen}{\textbf{+0.21\%}} \\
\cmidrule{2-8}
 & \multirow{3}{*}{Mat\'ern-$3/2$} & $\widehat{k}_\psi$  & 0.1434 & 0.2136 & 0.2498 & 0.2936 & 0.3239 \\
 &  & $k_0$  & 0.1325 & 0.2055 & 0.2497 & 0.2931 & 0.3257 \\
 &  & Rel.\ gain & \textcolor{gaingreen}{\textbf{+8.20\%}} & \textcolor{gaingreen}{\textbf{+3.94\%}} & \textcolor{gaingreen}{\textbf{+0.07\%}} & \textcolor{gaingreen}{\textbf{+0.17\%}} & \textcolor{lossred}{-0.56\%} \\
\cmidrule{2-8}
 & \multirow{3}{*}{Mat\'ern-$5/2$} & $\widehat{k}_\psi$  & 0.1453 & 0.2145 & 0.2529 & 0.2935 & 0.3251 \\
 &  & $k_0$  & 0.1334 & 0.2069 & 0.2502 & 0.2936 & 0.3257 \\
 &  & Rel.\ gain & \textcolor{gaingreen}{\textbf{+8.95\%}} & \textcolor{gaingreen}{\textbf{+3.69\%}} & \textcolor{gaingreen}{\textbf{+1.09\%}} & \textcolor{lossred}{-0.06\%} & \textcolor{lossred}{-0.17\%} \\
\cmidrule{2-8}
 & \multirow{3}{*}{RBF} & $\widehat{k}_\psi$  & 0.1519 & 0.2154 & 0.2494 & 0.2925 & 0.3265 \\
 &  & $k_0$  & 0.1356 & 0.2096 & 0.2517 & 0.2954 & 0.3222 \\
 &  & Rel.\ gain & \textcolor{gaingreen}{\textbf{+12.01\%}} & \textcolor{gaingreen}{\textbf{+2.81\%}} & \textcolor{lossred}{-0.90\%} & \textcolor{lossred}{-1.00\%} & \textcolor{gaingreen}{\textbf{+1.34\%}} \\
\midrule
\multirow{3}{*}{Qwen} & \multirow{3}{*}{Mat\'ern-$1/2$} & $\widehat{k}_\psi$  & 0.1231 & 0.1764 & 0.1986 & 0.2411 & 0.2732 \\
 &  & $k_0$  & 0.0969 & 0.1573 & 0.1955 & 0.2335 & 0.2745 \\
 &  & Rel.\ gain & \textcolor{gaingreen}{\textbf{+27.09\%}} & \textcolor{gaingreen}{\textbf{+12.14\%}} & \textcolor{gaingreen}{\textbf{+1.56\%}} & \textcolor{gaingreen}{\textbf{+3.23\%}} & \textcolor{lossred}{-0.47\%} \\
\cmidrule{2-8}
 & \multirow{3}{*}{Mat\'ern-$3/2$} & $\widehat{k}_\psi$  & 0.1191 & 0.1746 & 0.2002 & 0.2399 & 0.2719 \\
 &  & $k_0$  & 0.0890 & 0.1544 & 0.1963 & 0.2337 & 0.2727 \\
 &  & Rel.\ gain & \textcolor{gaingreen}{\textbf{+33.89\%}} & \textcolor{gaingreen}{\textbf{+13.11\%}} & \textcolor{gaingreen}{\textbf{+2.00\%}} & \textcolor{gaingreen}{\textbf{+2.68\%}} & \textcolor{lossred}{-0.30\%} \\
\cmidrule{2-8}
 & \multirow{3}{*}{Mat\'ern-$5/2$} & $\widehat{k}_\psi$  & 0.1138 & 0.1753 & 0.2016 & 0.2397 & 0.2716 \\
 &  & $k_0$  & 0.0867 & 0.1543 & 0.1964 & 0.2338 & 0.2716 \\
 &  & Rel.\ gain & \textcolor{gaingreen}{\textbf{+31.21\%}} & \textcolor{gaingreen}{\textbf{+13.63\%}} & \textcolor{gaingreen}{\textbf{+2.63\%}} & \textcolor{gaingreen}{\textbf{+2.52\%}} & 0.00\% \\
\cmidrule{2-8}
 & \multirow{3}{*}{RBF} & $\widehat{k}_\psi$  & 0.1094 & 0.1649 & 0.2028 & 0.2408 & 0.2714 \\
 &  & $k_0$  & 0.0796 & 0.1510 & 0.1954 & 0.2344 & 0.2703 \\
 &  & Rel.\ gain & \textcolor{gaingreen}{\textbf{+37.45\%}} & \textcolor{gaingreen}{\textbf{+9.23\%}} & \textcolor{gaingreen}{\textbf{+3.78\%}} & \textcolor{gaingreen}{\textbf{+2.73\%}} & \textcolor{gaingreen}{\textbf{+0.42\%}} \\
\bottomrule
\end{tabular}
\end{table}

\begin{table}[t]
\centering
\definecolor{gaingreen}{RGB}{0,128,55}
\definecolor{lossred}{RGB}{190,30,45}
\caption{AMAZON FASHION benchmark. Kernel ridge regression (KRR) with the learned kernel $\widehat{k}_\psi$ against KRR with the fixed reference kernel $k_0$ based on the test $R^2$ metric. The two share the same Lipschitz-controlled feature vector $\vz$ and the same reference kernel, and both select the ridge parameter based on the $N$ labeled points. The \emph{Rel.\ gain} (\%) row is the relative $R^2$ gain of KRR with $\hat{k}_{\psi}$ over KRR with $k_0$, as $\,(R^2_{\hat{k}_{\psi}}-R^2_{k_0})/R^2_{\mathrm{k_0}}$.}
\label{tab:q1amazonfashion}
\footnotesize
\setlength{\tabcolsep}{4pt}
\begin{tabular}{@{}lllccccc@{}}
\toprule
\textbf{Encoder} & \textbf{Kernel} & \textbf{Predictor} & $N=100$ & $N=200$ & $N=300$ & $N=500$ & $N=1000$ \\
\midrule
\multirow{3}{*}{BLIP-2} & \multirow{3}{*}{Mat\'ern-$1/2$} & $\widehat{k}_\psi$  & 0.1948 & 0.3042 & 0.3834 & 0.4342 & 0.4890 \\
 &  & $k_0$  & 0.1068 & 0.2687 & 0.3774 & 0.4232 & 0.4863 \\
 &  & Rel.\ gain & \textcolor{gaingreen}{\textbf{+82.31\%}} & \textcolor{gaingreen}{\textbf{+13.22\%}} & \textcolor{gaingreen}{\textbf{+1.60\%}} & \textcolor{gaingreen}{\textbf{+2.60\%}} & \textcolor{gaingreen}{\textbf{+0.56\%}} \\
\cmidrule{2-8}
 & \multirow{3}{*}{Mat\'ern-$3/2$} & $\widehat{k}_\psi$  & 0.1838 & 0.3033 & 0.3842 & 0.4334 & 0.4872 \\
 &  & $k_0$  & 0.0983 & 0.2693 & 0.3805 & 0.4274 & 0.4880 \\
 &  & Rel.\ gain & \textcolor{gaingreen}{\textbf{+86.92\%}} & \textcolor{gaingreen}{\textbf{+12.64\%}} & \textcolor{gaingreen}{\textbf{+0.98\%}} & \textcolor{gaingreen}{\textbf{+1.41\%}} & \textcolor{lossred}{-0.17\%} \\
\cmidrule{2-8}
 & \multirow{3}{*}{Mat\'ern-$5/2$} & $\widehat{k}_\psi$  & 0.1791 & 0.3017 & 0.3844 & 0.4335 & 0.4877 \\
 &  & $k_0$  & 0.0985 & 0.2705 & 0.3810 & 0.4292 & 0.4885 \\
 &  & Rel.\ gain & \textcolor{gaingreen}{\textbf{+81.77\%}} & \textcolor{gaingreen}{\textbf{+11.53\%}} & \textcolor{gaingreen}{\textbf{+0.91\%}} & \textcolor{gaingreen}{\textbf{+1.02\%}} & \textcolor{lossred}{-0.16\%} \\
\cmidrule{2-8}
 & \multirow{3}{*}{RBF} & $\widehat{k}_\psi$  & 0.1742 & 0.3000 & 0.3846 & 0.4330 & 0.4894 \\
 &  & $k_0$  & 0.0837 & 0.2603 & 0.3759 & 0.4244 & 0.4864 \\
 &  & Rel.\ gain & \textcolor{gaingreen}{\textbf{+108.06\%}} & \textcolor{gaingreen}{\textbf{+15.29\%}} & \textcolor{gaingreen}{\textbf{+2.30\%}} & \textcolor{gaingreen}{\textbf{+2.03\%}} & \textcolor{gaingreen}{\textbf{+0.62\%}} \\
\midrule
\multirow{3}{*}{LLaVA} & \multirow{3}{*}{Mat\'ern-$1/2$} & $\widehat{k}_\psi$  & 0.1098 & 0.1890 & 0.2805 & 0.3291 & 0.3977 \\
 &  & $k_0$  & 0.0760 & 0.1500 & 0.2754 & 0.3223 & 0.3949 \\
 &  & Rel.\ gain & \textcolor{gaingreen}{\textbf{+44.59\%}} & \textcolor{gaingreen}{\textbf{+26.00\%}} & \textcolor{gaingreen}{\textbf{+1.84\%}} & \textcolor{gaingreen}{\textbf{+2.12\%}} & \textcolor{gaingreen}{\textbf{+0.69\%}} \\
\cmidrule{2-8}
 & \multirow{3}{*}{Mat\'ern-$3/2$} & $\widehat{k}_\psi$  & 0.1037 & 0.1840 & 0.2781 & 0.3289 & 0.3953 \\
 &  & $k_0$  & 0.0682 & 0.1487 & 0.2766 & 0.3242 & 0.3953 \\
 &  & Rel.\ gain & \textcolor{gaingreen}{\textbf{+52.03\%}} & \textcolor{gaingreen}{\textbf{+23.71\%}} & \textcolor{gaingreen}{\textbf{+0.57\%}} & \textcolor{gaingreen}{\textbf{+1.43\%}} & 0.00\% \\
\cmidrule{2-8}
 & \multirow{3}{*}{Mat\'ern-$5/2$} & $\widehat{k}_\psi$  & 0.0987 & 0.1853 & 0.2782 & 0.3308 & 0.3953 \\
 &  & $k_0$  & 0.0688 & 0.1500 & 0.2770 & 0.3251 & 0.3954 \\
 &  & Rel.\ gain & \textcolor{gaingreen}{\textbf{+43.35\%}} & \textcolor{gaingreen}{\textbf{+23.49\%}} & \textcolor{gaingreen}{\textbf{+0.41\%}} & \textcolor{gaingreen}{\textbf{+1.76\%}} & \textcolor{lossred}{-0.04\%} \\
\cmidrule{2-8}
 & \multirow{3}{*}{RBF} & $\widehat{k}_\psi$  & 0.1002 & 0.1781 & 0.2808 & 0.3283 & 0.3956 \\
 &  & $k_0$  & 0.0599 & 0.1506 & 0.2788 & 0.3264 & 0.3956 \\
 &  & Rel.\ gain & \textcolor{gaingreen}{\textbf{+67.30\%}} & \textcolor{gaingreen}{\textbf{+18.28\%}} & \textcolor{gaingreen}{\textbf{+0.73\%}} & \textcolor{gaingreen}{\textbf{+0.60\%}} & \textcolor{gaingreen}{\textbf{+0.01\%}} \\
\midrule
\multirow{3}{*}{Qwen} & \multirow{3}{*}{Mat\'ern-$1/2$} & $\widehat{k}_\psi$  & 0.2573 & 0.3275 & 0.3912 & 0.4261 & 0.4779 \\
 &  & $k_0$  & 0.1980 & 0.2901 & 0.3795 & 0.4050 & 0.4641 \\
 &  & Rel.\ gain & \textcolor{gaingreen}{\textbf{+29.95\%}} & \textcolor{gaingreen}{\textbf{+12.90\%}} & \textcolor{gaingreen}{\textbf{+3.06\%}} & \textcolor{gaingreen}{\textbf{+5.20\%}} & \textcolor{gaingreen}{\textbf{+2.96\%}} \\
\cmidrule{2-8}
 & \multirow{3}{*}{Mat\'ern-$3/2$} & $\widehat{k}_\psi$  & 0.2456 & 0.3270 & 0.3900 & 0.4241 & 0.4778 \\
 &  & $k_0$  & 0.1875 & 0.2884 & 0.3828 & 0.4088 & 0.4668 \\
 &  & Rel.\ gain & \textcolor{gaingreen}{\textbf{+31.04\%}} & \textcolor{gaingreen}{\textbf{+13.38\%}} & \textcolor{gaingreen}{\textbf{+1.90\%}} & \textcolor{gaingreen}{\textbf{+3.75\%}} & \textcolor{gaingreen}{\textbf{+2.36\%}} \\
\cmidrule{2-8}
 & \multirow{3}{*}{Mat\'ern-$5/2$} & $\widehat{k}_\psi$  & 0.2487 & 0.3241 & 0.3888 & 0.4234 & 0.4773 \\
 &  & $k_0$  & 0.1840 & 0.2909 & 0.3834 & 0.4108 & 0.4682 \\
 &  & Rel.\ gain & \textcolor{gaingreen}{\textbf{+35.20\%}} & \textcolor{gaingreen}{\textbf{+11.41\%}} & \textcolor{gaingreen}{\textbf{+1.42\%}} & \textcolor{gaingreen}{\textbf{+3.05\%}} & \textcolor{gaingreen}{\textbf{+1.93\%}} \\
\cmidrule{2-8}
 & \multirow{3}{*}{RBF} & $\widehat{k}_\psi$  & 0.2475 & 0.3234 & 0.3878 & 0.4233 & 0.4794 \\
 &  & $k_0$  & 0.1720 & 0.2811 & 0.3812 & 0.4084 & 0.4690 \\
 &  & Rel.\ gain & \textcolor{gaingreen}{\textbf{+43.88\%}} & \textcolor{gaingreen}{\textbf{+15.04\%}} & \textcolor{gaingreen}{\textbf{+1.72\%}} & \textcolor{gaingreen}{\textbf{+3.66\%}} & \textcolor{gaingreen}{\textbf{+2.22\%}} \\
\bottomrule
\end{tabular}
\end{table}

\begin{table}[t]
\centering
\definecolor{gaingreen}{RGB}{0,128,55}
\definecolor{lossred}{RGB}{190,30,45}
\caption{QUECHUA-VALENCE benchmark. Kernel ridge regression (KRR) with the learned kernel $\widehat{k}_\psi$ against KRR with the fixed reference kernel $k_0$ based on the test $R^2$ metric. The two share the same Lipschitz-controlled feature vector $\vz$ and the same reference kernel, and both select the ridge parameter based on the $N$ labeled points. The \emph{Rel.\ gain} (\%) row is the relative $R^2$ gain of KRR with $\hat{k}_{\psi}$ over KRR with $k_0$, as $\,(R^2_{\hat{k}_{\psi}}-R^2_{k_0})/R^2_{\mathrm{k_0}}$.}
\label{tab:q1quechuaaudio}
\footnotesize
\setlength{\tabcolsep}{4pt}
\begin{tabular}{@{}lllccccc@{}}
\toprule
\textbf{Encoder} & \textbf{Kernel} & \textbf{Predictor} & $N=100$ & $N=200$ & $N=300$ & $N=500$ & $N=1000$ \\
\midrule
\multirow{3}{*}{Gemini~2} & \multirow{3}{*}{Mat\'ern-$1/2$} & $\widehat{k}_\psi$  & 0.0675 & 0.1758 & 0.2099 & 0.2331 & 0.2717 \\
 &  & $k_0$  & 0.0229 & 0.1620 & 0.2006 & 0.2267 & 0.2697 \\
 &  & Rel.\ gain & \textcolor{gaingreen}{\textbf{+194.89\%}} & \textcolor{gaingreen}{\textbf{+8.48\%}} & \textcolor{gaingreen}{\textbf{+4.60\%}} & \textcolor{gaingreen}{\textbf{+2.82\%}} & \textcolor{gaingreen}{\textbf{+0.72\%}} \\
\cmidrule{2-8}
 & \multirow{3}{*}{Mat\'ern-$3/2$} & $\widehat{k}_\psi$  & 0.0596 & 0.1753 & 0.2077 & 0.2336 & 0.2717 \\
 &  & $k_0$  & 0.0184 & 0.1618 & 0.2028 & 0.2275 & 0.2689 \\
 &  & Rel.\ gain & \textcolor{gaingreen}{\textbf{+223.91\%}} & \textcolor{gaingreen}{\textbf{+8.34\%}} & \textcolor{gaingreen}{\textbf{+2.41\%}} & \textcolor{gaingreen}{\textbf{+2.68\%}} & \textcolor{gaingreen}{\textbf{+1.03\%}} \\
\cmidrule{2-8}
 & \multirow{3}{*}{Mat\'ern-$5/2$} & $\widehat{k}_\psi$  & 0.0567 & 0.1743 & 0.2092 & 0.2325 & 0.2712 \\
 &  & $k_0$  & 0.0188 & 0.1620 & 0.2032 & 0.2277 & 0.2688 \\
 &  & Rel.\ gain & \textcolor{gaingreen}{\textbf{+201.60\%}} & \textcolor{gaingreen}{\textbf{+7.58\%}} & \textcolor{gaingreen}{\textbf{+2.95\%}} & \textcolor{gaingreen}{\textbf{+2.11\%}} & \textcolor{gaingreen}{\textbf{+0.90\%}} \\
\cmidrule{2-8}
 & \multirow{3}{*}{RBF} & $\widehat{k}_\psi$  & 0.0557 & 0.1712 & 0.2081 & 0.2320 & 0.2712 \\
 &  & $k_0$  & 0.0178 & 0.1497 & 0.1984 & 0.2224 & 0.2665 \\
 &  & Rel.\ gain & \textcolor{gaingreen}{\textbf{+212.92\%}} & \textcolor{gaingreen}{\textbf{+14.37\%}} & \textcolor{gaingreen}{\textbf{+4.90\%}} & \textcolor{gaingreen}{\textbf{+4.35\%}} & \textcolor{gaingreen}{\textbf{+1.76\%}} \\
\bottomrule
\end{tabular}
\end{table}

\begin{table}[t]
\centering
\definecolor{gaingreen}{RGB}{0,128,55}
\definecolor{lossred}{RGB}{190,30,45}
\caption{SPEECHOCEAN762-FLUENCY benchmark. Kernel ridge regression (KRR) with the learned kernel $\widehat{k}_\psi$ against KRR with the fixed reference kernel $k_0$ based on the test $R^2$ metric. The two share the same Lipschitz-controlled feature vector $\vz$ and the same reference kernel, and both select the ridge parameter based on the $N$ labeled points. The \emph{Rel.\ gain} (\%) row is the relative $R^2$ gain of KRR with $\hat{k}_{\psi}$ over KRR with $k_0$, as $\,(R^2_{\hat{k}_{\psi}}-R^2_{k_0})/R^2_{\mathrm{k_0}}$.}
\label{tab:q1speechoceanfluency}
\footnotesize
\setlength{\tabcolsep}{4pt}
\begin{tabular}{@{}lllccccc@{}}
\toprule
\textbf{Encoder} & \textbf{Kernel} & \textbf{Predictor} & $N=100$ & $N=200$ & $N=300$ & $N=500$ & $N=1000$ \\
\midrule
\multirow{3}{*}{Gemini~2} & \multirow{3}{*}{Mat\'ern-$1/2$} & $\widehat{k}_\psi$  & 0.2032 & 0.3310 & 0.3536 & 0.3901 & 0.4170 \\
 &  & $k_0$  & 0.1603 & 0.3123 & 0.3602 & 0.3747 & 0.4131 \\
 &  & Rel.\ gain & \textcolor{gaingreen}{\textbf{+26.78\%}} & \textcolor{gaingreen}{\textbf{+5.97\%}} & \textcolor{lossred}{-1.81\%} & \textcolor{gaingreen}{\textbf{+4.11\%}} & \textcolor{gaingreen}{\textbf{+0.94\%}} \\
\cmidrule{2-8}
 & \multirow{3}{*}{Mat\'ern-$3/2$} & $\widehat{k}_\psi$  & 0.2076 & 0.3314 & 0.3558 & 0.3915 & 0.4177 \\
 &  & $k_0$  & 0.1519 & 0.3105 & 0.3620 & 0.3794 & 0.4152 \\
 &  & Rel.\ gain & \textcolor{gaingreen}{\textbf{+36.65\%}} & \textcolor{gaingreen}{\textbf{+6.72\%}} & \textcolor{lossred}{-1.71\%} & \textcolor{gaingreen}{\textbf{+3.18\%}} & \textcolor{gaingreen}{\textbf{+0.60\%}} \\
\cmidrule{2-8}
 & \multirow{3}{*}{Mat\'ern-$5/2$} & $\widehat{k}_\psi$  & 0.1965 & 0.3276 & 0.3577 & 0.3926 & 0.4175 \\
 &  & $k_0$  & 0.1508 & 0.3096 & 0.3617 & 0.3791 & 0.4150 \\
 &  & Rel.\ gain & \textcolor{gaingreen}{\textbf{+30.32\%}} & \textcolor{gaingreen}{\textbf{+5.82\%}} & \textcolor{lossred}{-1.11\%} & \textcolor{gaingreen}{\textbf{+3.55\%}} & \textcolor{gaingreen}{\textbf{+0.59\%}} \\
\cmidrule{2-8}
 & \multirow{3}{*}{RBF} & $\widehat{k}_\psi$  & 0.1923 & 0.3294 & 0.3590 & 0.3910 & 0.4161 \\
 &  & $k_0$  & 0.1524 & 0.2840 & 0.3490 & 0.3457 & 0.3975 \\
 &  & Rel.\ gain & \textcolor{gaingreen}{\textbf{+26.20\%}} & \textcolor{gaingreen}{\textbf{+15.99\%}} & \textcolor{gaingreen}{\textbf{+2.86\%}} & \textcolor{gaingreen}{\textbf{+13.08\%}} & \textcolor{gaingreen}{\textbf{+4.69\%}} \\
\bottomrule
\end{tabular}
\end{table}

\begin{table}[t]
\centering
\definecolor{gaingreen}{RGB}{0,128,55}
\definecolor{lossred}{RGB}{190,30,45}
\caption{SPEECHOCEAN762-PROSODIC benchmark. Kernel ridge regression (KRR) with the learned kernel $\widehat{k}_\psi$ against KRR with the fixed reference kernel $k_0$ based on the test $R^2$ metric. The two share the same Lipschitz-controlled feature vector $\vz$ and the same reference kernel, and both select the ridge parameter based on the $N$ labeled points. The \emph{Rel.\ gain} (\%) row is the relative $R^2$ gain of KRR with $\hat{k}_{\psi}$ over KRR with $k_0$, as $\,(R^2_{\hat{k}_{\psi}}-R^2_{k_0})/R^2_{\mathrm{k_0}}$.}
\label{tab:q1speechoceanprosodic}
\footnotesize
\setlength{\tabcolsep}{4pt}
\begin{tabular}{@{}lllccccc@{}}
\toprule
\textbf{Encoder} & \textbf{Kernel} & \textbf{Predictor} & $N=100$ & $N=200$ & $N=300$ & $N=500$ & $N=1000$ \\
\midrule
\multirow{3}{*}{Gemini~2} & \multirow{3}{*}{Mat\'ern-$1/2$} & $\widehat{k}_\psi$  & 0.2165 & 0.3423 & 0.3487 & 0.3809 & 0.4128 \\
 &  & $k_0$  & 0.1868 & 0.3120 & 0.3463 & 0.3622 & 0.4077 \\
 &  & Rel.\ gain & \textcolor{gaingreen}{\textbf{+15.95\%}} & \textcolor{gaingreen}{\textbf{+9.70\%}} & \textcolor{gaingreen}{\textbf{+0.69\%}} & \textcolor{gaingreen}{\textbf{+5.16\%}} & \textcolor{gaingreen}{\textbf{+1.23\%}} \\
\cmidrule{2-8}
 & \multirow{3}{*}{Mat\'ern-$3/2$} & $\widehat{k}_\psi$  & 0.2099 & 0.3407 & 0.3503 & 0.3825 & 0.4127 \\
 &  & $k_0$  & 0.1783 & 0.3103 & 0.3482 & 0.3677 & 0.4099 \\
 &  & Rel.\ gain & \textcolor{gaingreen}{\textbf{+17.71\%}} & \textcolor{gaingreen}{\textbf{+9.81\%}} & \textcolor{gaingreen}{\textbf{+0.61\%}} & \textcolor{gaingreen}{\textbf{+4.02\%}} & \textcolor{gaingreen}{\textbf{+0.69\%}} \\
\cmidrule{2-8}
 & \multirow{3}{*}{Mat\'ern-$5/2$} & $\widehat{k}_\psi$  & 0.2144 & 0.3410 & 0.3508 & 0.3824 & 0.4133 \\
 &  & $k_0$  & 0.1769 & 0.3083 & 0.3473 & 0.3676 & 0.4096 \\
 &  & Rel.\ gain & \textcolor{gaingreen}{\textbf{+21.19\%}} & \textcolor{gaingreen}{\textbf{+10.61\%}} & \textcolor{gaingreen}{\textbf{+1.01\%}} & \textcolor{gaingreen}{\textbf{+4.02\%}} & \textcolor{gaingreen}{\textbf{+0.90\%}} \\
\cmidrule{2-8}
 & \multirow{3}{*}{RBF} & $\widehat{k}_\psi$  & 0.2144 & 0.3382 & 0.3564 & 0.3834 & 0.4098 \\
 &  & $k_0$  & 0.1818 & 0.2785 & 0.3313 & 0.3341 & 0.3934 \\
 &  & Rel.\ gain & \textcolor{gaingreen}{\textbf{+17.88\%}} & \textcolor{gaingreen}{\textbf{+21.42\%}} & \textcolor{gaingreen}{\textbf{+7.58\%}} & \textcolor{gaingreen}{\textbf{+14.75\%}} & \textcolor{gaingreen}{\textbf{+4.17\%}} \\
\bottomrule
\end{tabular}
\end{table}

\begin{table}[t]
\centering
\caption{\textbf{Q1} --- PAD-UFES-20-AGE benchmark. AdaKerNet using one of four distinct reference kernels, against the baselines on raw representations based on test $R^2$. Best per column in \textbf{bold}, second best \underline{underlined}.}
\label{tab:q2padufes20}
\footnotesize
\setlength{\tabcolsep}{4pt}
\begin{tabular}{@{}lccccc@{}}
\toprule
\textbf{Method} & $N=100$ & $N=200$ & $N=300$ & $N=500$ & $N=1000$ \\
\midrule
\multicolumn{6}{c}{\textbf{BLIP-2}} \\
\cmidrule(lr){1-6}
\multicolumn{6}{@{}l}{\textit{AdaKerNet, one row per reference kernel $k_0$}} \\
\textbf{AdaKerNet} (Mat\'ern-$1/2$) & \underline{0.2753} & 0.3226 & 0.3562 & \textbf{0.3930} & 0.4237 \\
\textbf{AdaKerNet} (Mat\'ern-$3/2$) & \textbf{0.2785} & \underline{0.3247} & \textbf{0.3579} & 0.3899 & \underline{0.4260} \\
\textbf{AdaKerNet} (Mat\'ern-$5/2$) & \textbf{0.2785} & 0.3229 & \underline{0.3567} & \underline{0.3903} & 0.4256 \\
\textbf{AdaKerNet} (RBF) & 0.2750 & \textbf{0.3251} & \textbf{0.3579} & 0.3891 & \textbf{0.4261} \\
\cmidrule(lr){1-6}
\multicolumn{6}{@{}l}{\textit{Baselines on raw representations $\ve$}} \\
MLP & 0.2586 & 0.2841 & 0.3359 & 0.3729 & 0.4129 \\
Transformer & 0.2121 & 0.2409 & 0.3009 & 0.3367 & 0.3757 \\
Autoencoder $+$ linear & -2.7412 & -0.7753 & -0.1098 & 0.1991 & 0.3462 \\
KRR & 0.1338 & 0.2386 & 0.2898 & 0.3467 & 0.4120 \\
SNGP & 0.1100 & 0.0894 & 0.1997 & 0.2601 & 0.3325 \\
\midrule
\multicolumn{6}{c}{\textbf{LLaVA}} \\
\cmidrule(lr){1-6}
\multicolumn{6}{@{}l}{\textit{AdaKerNet, one row per reference kernel $k_0$}} \\
\textbf{AdaKerNet} (Mat\'ern-$1/2$) & \textbf{0.3057} & \underline{0.3598} & \textbf{0.3978} & \textbf{0.4070} & 0.4466 \\
\textbf{AdaKerNet} (Mat\'ern-$3/2$) & 0.3023 & \textbf{0.3610} & \underline{0.3961} & 0.4051 & 0.4458 \\
\textbf{AdaKerNet} (Mat\'ern-$5/2$) & 0.3029 & 0.3578 & \underline{0.3961} & 0.4058 & \textbf{0.4477} \\
\textbf{AdaKerNet} (RBF) & \underline{0.3045} & 0.3588 & 0.3958 & \underline{0.4061} & \underline{0.4472} \\
\cmidrule(lr){1-6}
\multicolumn{6}{@{}l}{\textit{Baselines on raw representations $\ve$}} \\
MLP & 0.2615 & 0.2925 & 0.3399 & 0.3511 & 0.4082 \\
Transformer & 0.1883 & 0.1989 & 0.2648 & 0.2729 & 0.3701 \\
Autoencoder $+$ linear & -2.6360 & -0.8492 & -0.0904 & 0.1722 & 0.3371 \\
KRR & -0.1548 & 0.0680 & 0.1569 & 0.2577 & 0.3521 \\
SNGP & 0.0956 & 0.1409 & 0.1412 & 0.1887 & 0.2827 \\
\midrule
\multicolumn{6}{c}{\textbf{Qwen}} \\
\cmidrule(lr){1-6}
\multicolumn{6}{@{}l}{\textit{AdaKerNet, one row per reference kernel $k_0$}} \\
\textbf{AdaKerNet} (Mat\'ern-$1/2$) & 0.2722 & 0.3147 & 0.3516 & 0.3644 & 0.3915 \\
\textbf{AdaKerNet} (Mat\'ern-$3/2$) & 0.2760 & \underline{0.3158} & \underline{0.3525} & 0.3678 & \underline{0.3916} \\
\textbf{AdaKerNet} (Mat\'ern-$5/2$) & \textbf{0.2781} & 0.3144 & \textbf{0.3528} & \underline{0.3687} & 0.3908 \\
\textbf{AdaKerNet} (RBF) & \underline{0.2771} & \textbf{0.3167} & 0.3522 & \textbf{0.3690} & \textbf{0.3923} \\
\cmidrule(lr){1-6}
\multicolumn{6}{@{}l}{\textit{Baselines on raw representations $\ve$}} \\
MLP & -0.0157 & 0.0691 & 0.1214 & 0.1150 & 0.1604 \\
Transformer & -0.0741 & -0.0344 & -0.0162 & 0.0619 & 0.1699 \\
Autoencoder $+$ linear & -2.8743 & -1.1566 & -0.1991 & 0.0969 & 0.2710 \\
KRR & -0.1291 & 0.0347 & 0.1103 & 0.2012 & 0.2991 \\
SNGP & -0.0074 & 0.0040 & 0.0911 & 0.1527 & 0.1966 \\
\bottomrule
\end{tabular}
\end{table}

\begin{table}[t]
\centering
\caption{\textbf{Q1} --- SPEECHOCEAN762-FLUENCY benchmark. AdaKerNet using one of four distinct reference kernels, against the baselines on raw representations based on test $R^2$. Best per column in \textbf{bold}, second best \underline{underlined}.}
\label{tab:q2speechoceanfluency}
\footnotesize
\setlength{\tabcolsep}{4pt}
\begin{tabular}{@{}lccccc@{}}
\toprule
\textbf{Method} & $N=100$ & $N=200$ & $N=300$ & $N=500$ & $N=1000$ \\
\midrule
\multicolumn{6}{c}{\textbf{Gemini~2}} \\
\cmidrule(lr){1-6}
\multicolumn{6}{@{}l}{\textit{AdaKerNet, one row per reference kernel $k_0$}} \\
\textbf{AdaKerNet} (Mat\'ern-$1/2$) & \textbf{0.2168} & \textbf{0.3328} & 0.3519 & 0.3917 & 0.4194 \\
\textbf{AdaKerNet} (Mat\'ern-$3/2$) & \underline{0.2130} & 0.3316 & 0.3543 & 0.3946 & 0.4198 \\
\textbf{AdaKerNet} (Mat\'ern-$5/2$) & 0.2056 & 0.3307 & \underline{0.3564} & \underline{0.3960} & \underline{0.4199} \\
\textbf{AdaKerNet} (RBF) & 0.2033 & \underline{0.3321} & \textbf{0.3568} & \textbf{0.3963} & \textbf{0.4204} \\
\cmidrule(lr){1-6}
\multicolumn{6}{@{}l}{\textit{Baselines on raw representations $\ve$}} \\
MLP & 0.2092 & 0.2226 & 0.2454 & 0.2593 & 0.3611 \\
Transformer & 0.1973 & 0.1963 & 0.2077 & 0.2440 & 0.3250 \\
Autoencoder $+$ linear & -1.5754 & -0.6767 & -0.0158 & 0.1911 & 0.3039 \\
KRR & -0.2134 & 0.0791 & 0.1784 & 0.2829 & 0.3699 \\
SNGP & 0.1230 & 0.1356 & 0.1648 & 0.1513 & 0.2463 \\
\bottomrule
\end{tabular}
\end{table}

\begin{table}[t]
\centering
\caption{\textbf{Q2} --- PAD-UFES-20-AGE benchmark in the less sparse regime. AdaKerNet against the baselines on raw representations at every $N>1000$  AdaKerNet uses the RBF reference kernel reported in the main results. Best per column in \textbf{bold}, second best \underline{underlined}.}
\label{tab:q3padufes20}
\footnotesize
\setlength{\tabcolsep}{4pt}
\begin{tabular}{@{}lccc@{}}
\toprule
 & \multicolumn{1}{c}{BLIP-2} & \multicolumn{1}{c}{LLaVA} & \multicolumn{1}{c}{Qwen} \\
\cmidrule(lr){2-2}\cmidrule(lr){3-3}\cmidrule(lr){4-4}
\textbf{Method} & $N=1612$ & $N=1612$ & $N=1612$ \\
\midrule
\textbf{AdaKerNet} & 0.4365 & \textbf{0.4720} & \textbf{0.4106} \\
\cmidrule(lr){1-4}
\multicolumn{4}{@{}l}{\textit{Baselines on raw representations $\ve$}} \\
MLP & \textbf{0.4479} & \underline{0.4498} & 0.1642 \\
Transformer & 0.4054 & 0.4187 & 0.2243 \\
Autoencoder $+$ linear & 0.3870 & 0.3816 & 0.3209 \\
KRR & \underline{0.4476} & 0.4006 & \underline{0.3519} \\
SNGP & 0.3444 & 0.3593 & 0.2474 \\
\bottomrule
\end{tabular}
\end{table}

\begin{table}[t]
\centering
\caption{\textbf{Q2} --- PAD-UFES-20-MOLE-LESION in the less sparse regime. AdaKerNet against the baselines on raw representations at every $N>1000$. AdaKerNet uses the RBF reference kernel reported in the main results. Best per column in \textbf{bold}, second best \underline{underlined}.}
\label{tab:q3padufes20diameter}
\footnotesize
\setlength{\tabcolsep}{4pt}
\begin{tabular}{@{}lccc@{}}
\toprule
 & \multicolumn{1}{c}{BLIP-2} & \multicolumn{1}{c}{LLaVA} & \multicolumn{1}{c}{Qwen} \\
\cmidrule(lr){2-2}\cmidrule(lr){3-3}\cmidrule(lr){4-4}
\textbf{Method} & $N=1043$ & $N=1043$ & $N=1043$ \\
\midrule
\textbf{AdaKerNet} & \textbf{0.3716} & \textbf{0.3246} & \textbf{0.2724} \\
\cmidrule(lr){1-4}
\multicolumn{4}{@{}l}{\textit{Baselines on raw representations $\ve$}} \\
MLP & 0.1856 & 0.2134 & 0.0654 \\
Transformer & 0.2678 & 0.1260 & 0.1051 \\
Autoencoder $+$ linear & 0.3330 & 0.2478 & 0.1755 \\
KRR & \underline{0.3659} & \underline{0.2971} & \underline{0.2238} \\
SNGP & 0.3178 & 0.2262 & 0.1739 \\
\bottomrule
\end{tabular}
\end{table}

\begin{table}[t]
\centering
\caption{\textbf{Q2} --- AMAZON FASHION benchmark in the less sparse regime. AdaKerNet against the baselines on raw representations at every $N>1000$. AdaKerNet uses the Mat\'ern-$1/2$ reference kernel reported in the main results. Best per column in \textbf{bold}, second best \underline{underlined}.}
\label{tab:q3amazonfashion}
\footnotesize
\setlength{\tabcolsep}{4pt}
\begin{tabular}{@{}lcccccc@{}}
\toprule
 & \multicolumn{2}{c}{BLIP-2} & \multicolumn{2}{c}{LLaVA} & \multicolumn{2}{c}{Qwen} \\
\cmidrule(lr){2-3}\cmidrule(lr){4-5}\cmidrule(lr){6-7}
\textbf{Method} & $N=2000$ & $N=3486$ & $N=2000$ & $N=3486$ & $N=2000$ & $N=3486$ \\
\midrule
\textbf{AdaKerNet} & \textbf{0.5458} & \textbf{0.5792} & \textbf{0.4648} & \textbf{0.5019} & \textbf{0.5321} & \textbf{0.5608} \\
\cmidrule(lr){1-7}
\multicolumn{7}{@{}l}{\textit{Baselines on raw representations $\ve$}} \\
MLP & 0.4860 & \underline{0.5285} & 0.3132 & 0.3184 & 0.4307 & 0.4840 \\
Transformer & 0.4643 & 0.5108 & 0.3902 & 0.4202 & \underline{0.4596} & \underline{0.4993} \\
Autoencoder $+$ linear & 0.3388 & 0.3602 & 0.3154 & 0.3386 & 0.4096 & 0.4291 \\
KRR & \underline{0.4879} & 0.5281 & \underline{0.4012} & \underline{0.4495} & 0.4553 & 0.4953 \\
SNGP & 0.3773 & 0.4789 & 0.3298 & 0.3944 & 0.3439 & 0.4520 \\
\bottomrule
\end{tabular}
\end{table}

\begin{table}[t]
\centering
\caption{\textbf{Q2} --- QUECHUA-VALENCE benchmark in the less sparse regime. AdaKerNet against the baselines on raw representations at every $N>1000$. AdaKerNet uses the Mat\'ern-$5/2$ reference kernel reported in the main results. Best per column in \textbf{bold}, second best \underline{underlined}.}
\label{tab:q3quechuaaudio}
\footnotesize
\setlength{\tabcolsep}{4pt}
\begin{tabular}{@{}lcc@{}}
\toprule
\textbf{Method} & $N=2000$ & $N=2800$ \\
\midrule
\textbf{AdaKerNet} & \underline{0.3111} & \underline{0.3260} \\
\cmidrule(lr){1-3}
\multicolumn{3}{@{}l}{\textit{Baselines on raw representations $\ve$}} \\
MLP & 0.1416 & 0.2035 \\
Transformer & 0.1932 & 0.2167 \\
Autoencoder $+$ linear & 0.2520 & 0.2631 \\
KRR & \textbf{0.3236} & \textbf{0.3489} \\
SNGP & 0.2987 & 0.3147 \\
\bottomrule
\end{tabular}
\end{table}

\begin{table}[t]
\centering
\caption{\textbf{Q2} --- SPEECHOCEAN762-FLUENCY benchmark the less sparse regime. AdaKerNet against the baselines on raw representations at every $N>1000$. AdaKerNet uses the RBF reference kernel reported in the main results. Best per column in \textbf{bold}, second best \underline{underlined}.}
\label{tab:q3speechoceanfluency}
\footnotesize
\setlength{\tabcolsep}{4pt}
\begin{tabular}{@{}lc@{}}
\toprule
\textbf{Method} & $N=2000$ \\
\midrule
\textbf{AdaKerNet} & \textbf{0.4399} \\
\cmidrule(lr){1-2}
\multicolumn{2}{@{}l}{\textit{Baselines on raw representations $\ve$}} \\
MLP & 0.4079 \\
Transformer & 0.3722 \\
Autoencoder $+$ linear & 0.3454 \\
KRR & \underline{0.4299} \\
SNGP & 0.3756 \\
\bottomrule
\end{tabular}
\end{table}

\begin{table}[t]
\centering
\caption{\textbf{Q2} --- SPEECHOCEAN762-PROSODIC benchmark in the less sparse regime. AdaKerNet against the baselines on raw representations at every $N>1000$. AdaKerNet uses the Mat\'ern-$3/2$ reference kernel reported in the main results. Best per column in \textbf{bold}, second best \underline{underlined}.}
\label{tab:q3speechoceanprosodic}
\footnotesize
\setlength{\tabcolsep}{4pt}
\begin{tabular}{@{}lc@{}}
\toprule
\textbf{Method} & $N=2000$ \\
\midrule
\textbf{AdaKerNet} & \underline{0.4327} \\
\cmidrule(lr){1-2}
\multicolumn{2}{@{}l}{\textit{Baselines on raw representations $\ve$}} \\
MLP & 0.4085 \\
Transformer & 0.3850 \\
Autoencoder $+$ linear & 0.3622 \\
KRR & \textbf{0.4485} \\
SNGP & 0.3775 \\
\bottomrule
\end{tabular}
\end{table}

\begin{table}[t]
\centering
\caption{\textbf{Q3} --- AdaKerNet on the QUECHUA-VALENCE benchmark, Gemini~2 representation, evaluated for different values of $\lambda$, at each of four reference kernels based on the test $R^2$. For $\lambda \in \{0.1,1,10\}$ the value selected for that kernel and $N$ is given in parentheses. Best per column in \textbf{bold}.}
\label{tab:q4lambda}
\footnotesize
\setlength{\tabcolsep}{6pt}
\begin{tabular}{@{}lccc@{}}
\toprule
\multicolumn{4}{c}{\textbf{QUECHUA-VALENCE}} \\
\midrule
\textbf{Reference kernel} & $N=200$ & $N=300$ & $N=500$ \\
\midrule
\multicolumn{4}{@{}l}{\textit{$\lambda = 0.0001$}} \\
Mat\'ern-$1/2$ & 0.1721 & 0.2086 & 0.2297 \\
Mat\'ern-$3/2$ & 0.1720 & 0.2079 & 0.2320 \\
Mat\'ern-$5/2$ & 0.1705 & 0.2089 & 0.2330 \\
RBF & 0.1735 & 0.2087 & 0.2341 \\
\midrule
\multicolumn{4}{@{}l}{\textit{$\lambda \in \{0.1,\,1,\,10\}$}} \\
Mat\'ern-$1/2$ & 0.1692\,{\scriptsize(0.1)} & 0.2100\,{\scriptsize(0.1)} & 0.2333\,{\scriptsize(0.1)} \\
Mat\'ern-$3/2$ & 0.1749\,{\scriptsize(0.1)} & 0.2097\,{\scriptsize(0.1)} & \textbf{0.2350\,{\scriptsize(0.1)}} \\
Mat\'ern-$5/2$ & \textbf{0.1756\,{\scriptsize(0.1)}} & \textbf{0.2108\,{\scriptsize(0.1)}} & 0.2349\,{\scriptsize(0.1)} \\
RBF & 0.1755\,{\scriptsize(0.1)} & 0.2097\,{\scriptsize(0.1)} & 0.2338\,{\scriptsize(0.1)} \\
\midrule
\multicolumn{4}{@{}l}{\textit{$\lambda = 10000$}} \\
Mat\'ern-$1/2$ & 0.1407 & 0.1933 & 0.2189 \\
Mat\'ern-$3/2$ & 0.1487 & 0.1934 & 0.2225 \\
Mat\'ern-$5/2$ & 0.1440 & 0.1927 & 0.2249 \\
RBF & 0.1488 & 0.1973 & 0.2205 \\
\bottomrule
\end{tabular}
\end{table}

\begin{table}[!t]
\centering
\definecolor{gaingreen}{RGB}{0,128,55}
\definecolor{lossred}{RGB}{190,30,45}
\caption{\textbf AdaKerNet against kernel ridge regression on the raw representation $\ve$ with a Mat\'ern-$1/2$ kernel instead of the RBF kernel used in the main results, on all six benchmarks based on test $R^2$ metric. AdaKerNet uses the reference kernel reported for each benchmark in the main results: RBF for both PAD-UFES-20 targets and for SPEECHOCEAN762-FLUENCY, Mat\'ern-$1/2$ for AMAZON-FASHION, Mat\'ern-$5/2$ for QUECHUA-VALENCE and Mat\'ern-$3/2$ for SPEECHOCEAN762-PROSODIC. The higher entry at each $N$ is in \textbf{bold}. \emph{Err.\ red.}\ is AdaKerNet's relative reduction (\%) in test MSE over the KRR baseline as $(R^2_{\text{ours}}-R^2_{\text{KRR}})/(1-R^2_{\text{KRR}})$.}
\label{tab:q5}
\setlength{\tabcolsep}{7pt}
\setlength{\aboverulesep}{0pt}\setlength{\belowrulesep}{0pt}
\resizebox{\linewidth}{!}{%
\begin{tabular}{@{}l*{5}{c}@{\hspace{6pt}}|@{\hspace{6pt}}*{5}{c}@{}}
\toprule
\textbf{Method} & $N{=}100$ & $200$ & $300$ & $500$ & $1000$ & $N{=}100$ & $200$ & $300$ & $500$ & $1000$ \\
\midrule
\rule{0pt}{15pt} & & & \makebox[0pt]{\textbf{PAD-UFES-20-AGE / BLIP-2}} & & & & & \makebox[0pt]{\textbf{PAD-UFES-20-AGE / LLaVA}} & & \\[4pt]
KRR (Mat\'ern-$1/2$) on $\ve$ & 0.1263 & 0.2504 & 0.3025 & 0.3591 & 0.4168 & -0.0746 & 0.1445 & 0.2320 & 0.3244 & 0.4093 \\
AdaKerNet & \textbf{0.2750} & \textbf{0.3251} & \textbf{0.3579} & \textbf{0.3891} & \textbf{0.4261} & \textbf{0.3045} & \textbf{0.3588} & \textbf{0.3958} & \textbf{0.4061} & \textbf{0.4472} \\
Err.\ red. & \textcolor{gaingreen}{\textbf{$+$17.0\%}} & \textcolor{gaingreen}{\textbf{$+$10.0\%}} & \textcolor{gaingreen}{\textbf{$+$7.9\%}} & \textcolor{gaingreen}{\textbf{$+$4.7\%}} & \textcolor{gaingreen}{\textbf{$+$1.6\%}} & \textcolor{gaingreen}{\textbf{$+$35.3\%}} & \textcolor{gaingreen}{\textbf{$+$25.0\%}} & \textcolor{gaingreen}{\textbf{$+$21.3\%}} & \textcolor{gaingreen}{\textbf{$+$12.1\%}} & \textcolor{gaingreen}{\textbf{$+$6.4\%}} \\[2pt]
\midrule
\rule{0pt}{15pt} & & & \makebox[0pt]{\textbf{PAD-UFES-20-AGE / Qwen}} & & & & & \makebox[0pt]{\textbf{PAD-UFES-20-MOLE-LESION / BLIP-2}} & & \\[4pt]
KRR (Mat\'ern-$1/2$) on $\ve$ & -0.0611 & 0.1121 & 0.1909 & 0.2813 & 0.3675 & -0.0323 & 0.1082 & 0.1667 & 0.2624 & 0.3380 \\
AdaKerNet & \textbf{0.2771} & \textbf{0.3167} & \textbf{0.3522} & \textbf{0.3690} & \textbf{0.3923} & \textbf{0.1777} & \textbf{0.2291} & \textbf{0.2731} & \textbf{0.3347} & \textbf{0.3716} \\
Err.\ red. & \textcolor{gaingreen}{\textbf{$+$31.9\%}} & \textcolor{gaingreen}{\textbf{$+$23.0\%}} & \textcolor{gaingreen}{\textbf{$+$19.9\%}} & \textcolor{gaingreen}{\textbf{$+$12.2\%}} & \textcolor{gaingreen}{\textbf{$+$3.9\%}} & \textcolor{gaingreen}{\textbf{$+$20.3\%}} & \textcolor{gaingreen}{\textbf{$+$13.6\%}} & \textcolor{gaingreen}{\textbf{$+$12.8\%}} & \textcolor{gaingreen}{\textbf{$+$9.8\%}} & \textcolor{gaingreen}{\textbf{$+$5.1\%}} \\[2pt]
\midrule
\rule{0pt}{15pt} & & & \makebox[0pt]{\textbf{PAD-UFES-20-MOLE-LESION / LLaVA}} & & & & & \makebox[0pt]{\textbf{PAD-UFES-20-MOLE-LESION / Qwen}} & & \\[4pt]
KRR (Mat\'ern-$1/2$) on $\ve$ & -0.2055 & 0.0132 & 0.0932 & 0.2029 & 0.2888 & -0.0404 & 0.0578 & 0.1040 & 0.1816 & 0.2588 \\
AdaKerNet & \textbf{0.1475} & \textbf{0.2134} & \textbf{0.2497} & \textbf{0.2925} & \textbf{0.3246} & \textbf{0.1356} & \textbf{0.1731} & \textbf{0.1987} & \textbf{0.2380} & \textbf{0.2724} \\
Err.\ red. & \textcolor{gaingreen}{\textbf{$+$29.3\%}} & \textcolor{gaingreen}{\textbf{$+$20.3\%}} & \textcolor{gaingreen}{\textbf{$+$17.3\%}} & \textcolor{gaingreen}{\textbf{$+$11.2\%}} & \textcolor{gaingreen}{\textbf{$+$5.0\%}} & \textcolor{gaingreen}{\textbf{$+$16.9\%}} & \textcolor{gaingreen}{\textbf{$+$12.2\%}} & \textcolor{gaingreen}{\textbf{$+$10.6\%}} & \textcolor{gaingreen}{\textbf{$+$6.9\%}} & \textcolor{gaingreen}{\textbf{$+$1.8\%}} \\[2pt]
\midrule
\rule{0pt}{15pt} & & & \makebox[0pt]{\textbf{AMAZON-FASHION / BLIP-2}} & & & & & \makebox[0pt]{\textbf{AMAZON-FASHION / LLaVA}} & & \\[4pt]
KRR (Mat\'ern-$1/2$) on $\ve$ & -0.0577 & 0.1151 & 0.1927 & 0.2736 & 0.3747 & -0.1145 & 0.0514 & 0.1256 & 0.1988 & 0.2938 \\
AdaKerNet & \textbf{0.2077} & \textbf{0.3105} & \textbf{0.3856} & \textbf{0.4349} & \textbf{0.4888} & \textbf{0.1206} & \textbf{0.1916} & \textbf{0.2812} & \textbf{0.3318} & \textbf{0.3957} \\
Err.\ red. & \textcolor{gaingreen}{\textbf{$+$25.1\%}} & \textcolor{gaingreen}{\textbf{$+$22.1\%}} & \textcolor{gaingreen}{\textbf{$+$23.9\%}} & \textcolor{gaingreen}{\textbf{$+$22.2\%}} & \textcolor{gaingreen}{\textbf{$+$18.2\%}} & \textcolor{gaingreen}{\textbf{$+$21.1\%}} & \textcolor{gaingreen}{\textbf{$+$14.8\%}} & \textcolor{gaingreen}{\textbf{$+$17.8\%}} & \textcolor{gaingreen}{\textbf{$+$16.6\%}} & \textcolor{gaingreen}{\textbf{$+$14.4\%}} \\[2pt]
\midrule
\rule{0pt}{15pt} & & & \makebox[0pt]{\textbf{AMAZON-FASHION / Qwen}} & & & & & \makebox[0pt]{\textbf{QUECHUA-VALENCE / Gemini~2}} & & \\[4pt]
KRR (Mat\'ern-$1/2$) on $\ve$ & -0.1005 & 0.0877 & 0.1815 & 0.2772 & 0.3871 & -0.0540 & 0.0638 & 0.1105 & 0.1653 & 0.2378 \\
AdaKerNet & \textbf{0.2618} & \textbf{0.3337} & \textbf{0.3920} & \textbf{0.4261} & \textbf{0.4782} & \textbf{0.0620} & \textbf{0.1756} & \textbf{0.2108} & \textbf{0.2349} & \textbf{0.2719} \\
Err.\ red. & \textcolor{gaingreen}{\textbf{$+$32.9\%}} & \textcolor{gaingreen}{\textbf{$+$27.0\%}} & \textcolor{gaingreen}{\textbf{$+$25.7\%}} & \textcolor{gaingreen}{\textbf{$+$20.6\%}} & \textcolor{gaingreen}{\textbf{$+$14.9\%}} & \textcolor{gaingreen}{\textbf{$+$11.0\%}} & \textcolor{gaingreen}{\textbf{$+$11.9\%}} & \textcolor{gaingreen}{\textbf{$+$11.3\%}} & \textcolor{gaingreen}{\textbf{$+$8.3\%}} & \textcolor{gaingreen}{\textbf{$+$4.5\%}} \\[2pt]
\midrule
\rule{0pt}{15pt} & & & \makebox[0pt]{\textbf{SPEECHOCEAN762-FLUENCY / Gemini~2}} & & & & & \makebox[0pt]{\textbf{SPEECHOCEAN762-PROSODIC / Gemini~2}} & & \\[4pt]
KRR (Mat\'ern-$1/2$) on $\ve$ & -0.3622 & -0.0176 & 0.1004 & 0.2272 & 0.3381 & -0.1612 & 0.0851 & 0.1730 & 0.2655 & 0.3569 \\
AdaKerNet & \textbf{0.2033} & \textbf{0.3321} & \textbf{0.3568} & \textbf{0.3963} & \textbf{0.4204} & \textbf{0.2230} & \textbf{0.3437} & \textbf{0.3515} & \textbf{0.3873} & \textbf{0.4155} \\
Err.\ red. & \textcolor{gaingreen}{\textbf{$+$41.5\%}} & \textcolor{gaingreen}{\textbf{$+$34.4\%}} & \textcolor{gaingreen}{\textbf{$+$28.5\%}} & \textcolor{gaingreen}{\textbf{$+$21.9\%}} & \textcolor{gaingreen}{\textbf{$+$12.4\%}} & \textcolor{gaingreen}{\textbf{$+$33.1\%}} & \textcolor{gaingreen}{\textbf{$+$28.3\%}} & \textcolor{gaingreen}{\textbf{$+$21.6\%}} & \textcolor{gaingreen}{\textbf{$+$16.6\%}} & \textcolor{gaingreen}{\textbf{$+$9.1\%}} \\
\bottomrule
\end{tabular}%
}
\end{table}

\begin{table}[!t]
\centering
\definecolor{gaingreen}{RGB}{0,128,55}
\definecolor{lossred}{RGB}{190,30,45}
\caption{Task-adaptive kernel deformation stage of AdaKerNet against \emph{spectral} kernel learning on the same Lipschitz-controlled features $\vz$, assessed using the test $R^2$ metric. \textsc{TRF on} $\vz$ is Tuned Random Features \citep{shilton2022trf}: frequencies are drawn once from the reference density and a per-frequency density ratio is learned alongside the weights. AdaKerNet instead deforms the reference kernel through a learned feature map, which need not remain translation-invariant. On each benchmark both are reported at that benchmark's own reference kernel, which is the density TRF samples its frequencies from, so neither is given a choice the other lacks. The \emph{Rel.\ gain} row gives AdaKerNet's relative $R^2$ gain (\%) over TRF as $(R^2_{\mathrm{AdaKerNet}}-R^2_{\mathrm{TRF}})/R^2_{\mathrm{TRF}}$.}
\label{tab:trfz_vs_adakernet}
\setlength{\tabcolsep}{7pt}
\setlength{\aboverulesep}{0pt}\setlength{\belowrulesep}{0pt}
\resizebox{\linewidth}{!}{%
\begin{tabular}{@{}l*{5}{c}@{\hspace{6pt}}|@{\hspace{6pt}}*{5}{c}@{}}
\toprule
\textbf{Method} & $N{=}100$ & $200$ & $300$ & $500$ & $1000$ & $N{=}100$ & $200$ & $300$ & $500$ & $1000$ \\
\midrule
\rule{0pt}{15pt} & & & \makebox[0pt]{\textbf{PAD-UFES-20-MOLE-LESION / BLIP-2}} & & & & & \makebox[0pt]{\textbf{PAD-UFES-20-MOLE-LESION / LLaVA}} & & \\[4pt]
TRF on $\vz$ & 0.1513 & 0.2255 & 0.2713 & 0.3395 & 0.3815 & 0.1252 & 0.2029 & 0.2413 & 0.2865 & 0.3143 \\
AdaKerNet & 0.1777 & 0.2291 & 0.2731 & 0.3347 & 0.3716 & 0.1475 & 0.2134 & 0.2497 & 0.2925 & 0.3246 \\
\quad Rel.\ gain & \textcolor{gaingreen}{\textbf{$+$17.4\%}} & \textcolor{gaingreen}{\textbf{$+$1.6\%}} & \textcolor{gaingreen}{\textbf{$+$0.6\%}} & \textcolor{lossred}{$-$1.4\%} & \textcolor{lossred}{$-$2.6\%} & \textcolor{gaingreen}{\textbf{$+$17.8\%}} & \textcolor{gaingreen}{\textbf{$+$5.2\%}} & \textcolor{gaingreen}{\textbf{$+$3.4\%}} & \textcolor{gaingreen}{\textbf{$+$2.1\%}} & \textcolor{gaingreen}{\textbf{$+$3.3\%}} \\[2pt]
\midrule
\rule{0pt}{15pt} & & & \makebox[0pt]{\textbf{PAD-UFES-20-MOLE-LESION / Qwen}} & & & & & \makebox[0pt]{\textbf{QUECHUA-VALENCE / Gemini~2}} & & \\[4pt]
TRF on $\vz$ & 0.1100 & 0.1679 & 0.1935 & 0.2331 & 0.2760 & 0.0261 & 0.1737 & 0.2073 & 0.2338 & 0.2753 \\
AdaKerNet & 0.1356 & 0.1731 & 0.1987 & 0.2380 & 0.2724 & 0.0620 & 0.1756 & 0.2108 & 0.2349 & 0.2719 \\
\quad Rel.\ gain & \textcolor{gaingreen}{\textbf{$+$23.2\%}} & \textcolor{gaingreen}{\textbf{$+$3.1\%}} & \textcolor{gaingreen}{\textbf{$+$2.7\%}} & \textcolor{gaingreen}{\textbf{$+$2.1\%}} & \textcolor{lossred}{$-$1.3\%} & \textcolor{gaingreen}{\textbf{$+$137.7\%}} & \textcolor{gaingreen}{\textbf{$+$1.1\%}} & \textcolor{gaingreen}{\textbf{$+$1.7\%}} & \textcolor{gaingreen}{\textbf{$+$0.4\%}} & \textcolor{lossred}{$-$1.2\%} \\[2pt]
\midrule
\rule{0pt}{15pt} & & & \makebox[0pt]{\textbf{SPEECHOCEAN762-FLUENCY / Gemini~2}} & & & & & \makebox[0pt]{\textbf{SPEECHOCEAN762-PROSODIC / Gemini~2}} & & \\[4pt]
TRF on $\vz$ & 0.1551 & 0.3295 & 0.3565 & 0.3893 & 0.4103 & 0.1861 & 0.3403 & 0.3511 & 0.3806 & 0.4031 \\
AdaKerNet & 0.2033 & 0.3321 & 0.3568 & 0.3963 & 0.4204 & 0.2230 & 0.3437 & 0.3515 & 0.3873 & 0.4155 \\
\quad Rel.\ gain & \textcolor{gaingreen}{\textbf{$+$31.0\%}} & \textcolor{gaingreen}{\textbf{$+$0.8\%}} & \textcolor{gaingreen}{\textbf{$+$0.1\%}} & \textcolor{gaingreen}{\textbf{$+$1.8\%}} & \textcolor{gaingreen}{\textbf{$+$2.5\%}} & \textcolor{gaingreen}{\textbf{$+$19.8\%}} & \textcolor{gaingreen}{\textbf{$+$1.0\%}} & \textcolor{gaingreen}{\textbf{$+$0.1\%}} & \textcolor{gaingreen}{\textbf{$+$1.8\%}} & \textcolor{gaingreen}{\textbf{$+$3.1\%}} \\
\bottomrule
\end{tabular}%
}
\end{table}

\section{Constrained Interpretation of AdaKerNet and Remarks}

\paragraph{Constrained interpretation.} Let the learned features be stacked row-wise as  $   \Phi_\psi= [\vphi_\psi(\vz_1),\dots,\vphi_\psi(\vz_N)]^{\!\top}
    \in\R^{N\times F}$,  the reference kernel matrix defined as $   [K_0]_{ij}=k_0(\vz_i,\vz_j)$,
and let $H_\omega(\Phi_\psi)$ denote row-wise application of the MLP as $H_\omega(\Phi_\psi)
    =
    \begin{bmatrix}
    h_\omega(\vphi_\psi(\vz_1)) & \ldots &
    h_\omega(\vphi_\psi(\vz_N))
    \end{bmatrix}^{\!\top}$.
Then, the regression objective can be written as
\begin{equation}
    \mathcal L(\psi,\omega)
    =
    \frac{\lambda}{N}
    \norm{H_\omega(\Phi_\psi)-\bm{y}}_2^2
    +
    \frac{1}{N^2}
    \norm{\widehat K_\psi-K_0}_F^2
\label{eq:loss-matrix}
\end{equation}
with $\widehat K_\psi=\Phi_\psi\Phi_\psi^{\!\top} \succeq0 $. Equivalently, define the Gram-matrix deviation
\begin{equation}
    D_\psi:=\widehat K_\psi-K_0.
\end{equation}
Then the kernel regularizer is simply $\mathcal L_{\mathrm{ker}} = \norm{D_\psi}_F^2/N^2$,
which explicitly penalizes the average squared deviation of the learned pairwise similarities from those of the reference kernel.
The penalized objective in \eqref{eq:loss-matrix} is the Lagrangian form
of the constrained task-adaptation problem
\begin{equation}
\begin{aligned}
    \min_{\psi,\omega}\quad
    & \frac{1}{N^2}\norm{\widehat K_\psi-K_0}_F^2,\\
    \text{subject to}\quad
    &\frac{1}{N} \norm{H_\omega(\Phi_\psi)-\bm{y}}_2^2
      \leq\varepsilon.
    \label{eq:constrained-objective}
\end{aligned}
\end{equation}
This constrained formulation seeks a regularized feature map whose induced kernel does not significantly deviate from the reference kernel while enabling the neural predictor to satisfy a prescribed training-error tolerance. It interprets task-adaptive kernel deformation as balancing preservation of the reference similarity structure with the predictive requirements of the
downstream task. The two formulations are related
through the Lagrange multiplier $\lambda$, although the correspondence
between $\lambda$ and $\varepsilon$ is generally implicit.

\paragraph{Remark 1 (Why the learned kernel is task-adaptive).}
Although $k_0$ is fixed, the effective kernel $\widehat{k}_\psi$ changes
during training because $\psi$ receives gradients from both losses:
\begin{equation}
    \nabla_\psi\mathcal L
    =
    \lambda \nabla_\psi\mathcal L_{\mathrm{pred}}
    +\nabla_\psi\mathcal L_{\mathrm{ker}}.
    \label{eq:joint-gradient}
\end{equation}
The prediction gradient encourages changes to $\widehat{k}_\psi$ that
improve the downstream task, whereas the kernel gradient discourages
large deviations from $k_0$. Consequently, $k_0$ acts as a soft
structural prior rather than as an immutable similarity function.

\noindent The role of $\lambda$ can be summarized as follows:
\begin{itemize}
    \item For $\lambda \approx 0$, learning is governed mostly by kernel reconstruction. The feature map $\vphi_\psi$ receives minimal task-specific supervision through the prediction loss.

    \item For finite $\lambda>0$, both objectives shape the learned features: kernel reconstruction anchors their induced similarities to the reference kernel, while
    supervised prediction encourages task-adaptive deviations. Increasing $\lambda$ places greater emphasis on predictive fit relative to reference-kernel reconstruction fidelity.

    \item As $\lambda\rightarrow\infty$, the prediction loss
    dominates and the kernel reconstruction becomes negligible relative to prediction, approaching purely supervised learning with neural prediction MLP head.
\end{itemize}

\paragraph{Remark 2 (Relation to kernel ridge regression).}
Kernel ridge regression (KRR) with the fixed kernel $k_0$ has the predictor
\begin{equation}
    f_{\mathrm{KRR}}(\vz)
    =\sum_{m=1}^{M}\alpha_m k_0(\vz,\vz_m),
    \qquad
    \bm{\alpha}=(K_0+\eta I)^{-1}\bm{y},
    \label{eq:krr}
\end{equation}
up to the convention used to scale the ridge parameter $\eta$. KRR
optimizes the sample-dependent coefficients $\bm{\alpha}$ while keeping
$k_0$ fixed. In contrast, our method learns the feature map defining
$\widehat{k}_\psi$ jointly with the nonlinear predictor $h_\omega$.
Accordingly, the MLP predictor is not, in general, a representer-theorem
expansion of the form
$\sum_m\alpha_m\widehat{k}_\psi(\vz,\vz_m)$; rather, it is a nonlinear
predictor operating on the explicit features of the learned kernel.

\section{Additional experimental details}

\subsection{Dataset description}

\textbf{PAD-UFES-20-AGE.} This dataset consists of dermatological smartphone images of
skin lesions, each accompanied by a free-text clinical description and a
patient questionnaire. The target variable to be predicted is the patient age. The tabular block contributes 13 standardized clinical
variables. From the available lesions, 367 are reserved for testing, and 1,612 are used as the labeled pool that can be used for training (and where applicable, validation); because several lesions may belong to one patient, the split is not patient-disjoint.

\textbf{PAD-UFES-20-MOLE-LESION.} The same image, text and
questionnaire sources, with the logarithm of the larger recorded lesion
diameter in millimetres as the target. Only the lesions for which
both diameters are recorded are retained (the remaining 808 are dropped
rather than imputed). Both diameter fields are removed from the tabular
block, as is the indicator of whether the lesion was biopsied, and
patient age is added in their place. The split
is patient-disjoint: 1{,}043 is the available labeled pool, with 284 being reserved for testing, with no patient appearing on both sides.

\textbf{AMAZON FASHION.} This dataset consists of product listings comprising a photograph, review and description text, and structured catalogue attributes. We predict the
logarithm of the product price. The tabular block contributes 174
dimensions after standardization and one-hot encoding of categorical
fields. 3,486 are the available labeled data and 747 are the test ones.

\textbf{QUECHUA-VALENCE.} This dataset comprises spoken utterances in Quechua Collao with
aligned transcripts.
Four native-speaker annotators rated each utterance for valence on a
five-point scale; the target is the mean rating rescaled to $[0,1]$. We use a random subsample of 2{,}800 utterances available for training and 600 are reserved for testing.

\textbf{SPEECHOCEAN762-FLUENCY.} Five thousand English utterances from
250 non-native speakers, 20 utterances each, with 4{,}947 distinct
transcripts. Each utterance carries expert pronunciation
ratings; we predict the fluency score. We use the corpus's own official
split rather than the random partition used elsewhere: 2{,}000 available for training,
and 2{,}500 test utterances, speaker-disjoint with no
speaker appearing on both sides.

\textbf{SPEECHOCEAN762-PROSODIC.} The same utterances, and the same speaker-disjoint official split as
above, with the prosodic score as the target.

\subsection{Representation extraction from MLLMs}

For PAD-UFES-20-AGE, PAD-UFES-20-MOLE-LESION and AMAZON FASHION, we utilize the visual representation
from the MLLM: the Q-Former output for BLIP-2, the projected patch
features for LLaVA-1.5, the post-merger visual tokens for Qwen2.5-VL
and concatenate it with text embeddings for each model and the standardized tabular attributes. 
The resultant representations are of 1{,}549, 4{,}877 and 4{,}365 dimensions on
PAD-UFES-20-AGE, 1{,}547, 4{,}875 and 4{,}363 on PAD-UFES-20-MOLE-LESION, and 1{,}710, 5{,}038 and 4{,}526 on AMAZON FASHION.

For QUECHUA-VALENCE and both SPEECHOCEAN762-FLUENCY and SPEECHOCEAN762-PROSODIC the Gemini
Embedding 2 MLLM, processes the audio and its transcript jointly and returns a single fused 768-dimensional representation; thus no per-modality decomposition exists. 

We would like to note that 
where a baseline requires a token sequence (the transformer decoder of B-II), for PAD-UFES-20 (AGE and MOLE-LESION) and AMAZON FASHION, the three tokens correspond to the image, text, and tabular modalities, allowing attention to operate directly across modalities. For QUECHUA-VALENCE, SPEECHOCEAN762-FLUENCY and SPEECHOCEAN762-PROSODIC however, the representation is already internally fused and does not admit a natural modality-wise decomposition. We therefore partition $\ve$ into three equal contiguous segments to preserve the same architecture across benchmarks. These segments are purely structural and do not correspond to distinct modalities; consequently, the resulting attention weights should not be interpreted as modality-specific.

Representations are extracted once per benchmark and MLLM encoder arm and held frozen throughout. No MLLM is fine-tuned, and no gradient reaches the
MLLM encoder at any point in training.

\subsection{Detailed AdaKerNet implementation and training}

\textbf{Lipschitz-controlled feature extraction block.} The frozen representation $\ve$ is mapped to $\vz$ by a two-layer network,
$\ve \mapsto 512 \mapsto 256$, with GELU activations and dropout $0.25$
after each layer. Both linear layers are spectrally normalized, which
bounds the Lipschitz constant of the map and is what makes 
$\vz$ Lipschitz-controlled.  Supervision is provided by a
kernel ridge head applied to $\vz$ with the ridge parameter being fixed at $1$; gradients are propagated through the differentiable KRR solve into the spectrally normalized neural feature extractor, allowing the labeled supervision and reference kernel within KRR to directly shape the learned representation $\vz$.  The head is
discarded after this stage and $\vz$ is frozen. 

\textbf{Task-adaptive neural kernel representation block.} The map
$\vphi_\psi(\cdot)$ is a three-layer MLP,
$\vz \mapsto 512 \mapsto 256 \mapsto 128$, with ReLU activations, whose
output $\vrho$ induces the learned kernel
$\widehat{k}_\psi = \vrho\vrho^\top$. The prediction head
$h_\omega(\cdot)$ is a two-layer MLP, $\vrho \mapsto 256 \mapsto 1$,
also with ReLU.

\textit{The reference kernel.} $k_0$ is computed once on the frozen
$\vz$, with its length scale computed using the available labeled data. In the main results and ablation studies, we assess AdaKerNet's performance considering four distinct kernels types; namely Mat\'ern-$1/2$, Mat\'ern-$3/2$, Mat\'ern-$5/2$ and RBF.

\textit{Optimization.} The Lipschitz-controlled feature extraction block uses AdamW with learning rate
$5\times 10^{-4}$, weight decay $10^{-3}$ and a cosine schedule over 500
epochs, with gradients clipped to norm $1$. The task-adaptive neural kernel representation block uses Adam with learning rate $10^{-3}$ over 500 epochs, reducing the learning rate by
half after 20 epochs without improvement.
Both use minibatches of 256.

\textit{Selection of $\lambda$.} For all benchmarks, reference kernels, and label budgets $N$, we select $\lambda \in \{0.1,1,10\}$ using the available labeled data. We note that considering a broader set of candidate $\lambda$ values may yield further performance improvements beyond those reported in the main results and ablation studies.

\section{Ablation Studies}
\label{sec:supp_ablation_studies}

We provide additional experiments to examine the contributions of AdaKerNet's components and its behavior under different modeling choices and levels of supervision. We first evaluate the significance of Lipschitz-controlled features and the joint learning of the kernel representation and neural predictor. We then investigate the sensitivity to the reference-kernel choice, performance at larger label budgets, AdaKerNet's performance under different values of $\lambda$, alternative kernels for the KRR baseline. Unless otherwise stated, results follow the experimental setup described in Section~2 and report mean test $R^2$ over ten runs with different labeled subsets and initializations.

\subsection{Benefits of Lipschitz-controlled features}
\label{sec:supp_distance_features}

Table~3 examines whether alternatively using the learned Lipschitz-controlled features $\vz$ instead of the frozen MLLM representations $\ve$ improves downstream prediction performance. We consider two predictors using $\vz$: an MLP with the same downstream architecture apart from its input dimensionality, and a KRR head using the same RBF kernel form compared to the baselines B-I and B-IV respectively. The evaluated settings comprise both PAD-UFES-20 targets with three MLLM encoders and SPEECHOCEAN762-FLUENCY with Gemini Embedding~2, each at five label budgets, giving 35 configurations per predictor.

Using $\vz$ improves the MLP score in 33 of 35 configurations and the KRR score in 34 of 35. The corresponding relative reductions in test MSE reach $28.5\%$ for MLP and $38.7\%$ for KRR. For example, on PAD-UFES-20-AGE with Qwen2.5-VL and $N=200$, the MLP score increases from $0.0691$ to $0.3196$, while the KRR score increases from $0.0347$ to $0.3052$. These changes correspond to MSE reductions of $26.9\%$ and $28.0\%$, respectively. On the same task with LLaVA-1.5 and $N=100$, KRR improves from $-0.1548$ on $\ve$ to $0.2924$ on $\vz$, illustrating that the learned features can make a fixed kernel form substantially more useful for prediction.

The exceptions are limited. For MLP, using $\vz$ slightly reduces $R^2$ on PAD-UFES-20-AGE with BLIP-2 at $N=500$ and on SPEECHOCEAN762-FLUENCY at $N=100$. For KRR, the only reduction occurs on PAD-UFES-20-MOLE-LESION with BLIP-2 at the largest reported budget. Thus, Lipschitz-controlled feature learning is beneficial in most evaluated settings, and the learned features support both neural and kernel-based prediction.

\subsection{Benefits of joint neural predictor and kernel representation learning}
\label{sec:supp_joint_learning}

\paragraph{Complete decoder versus MLP on $\vz$.}
Table~4 assesses whether AdaKerNet as a whole end-to-end pipeline provides additional benefits over a neural MLP predictor once the Lipschitz-controlled features have been obtained. Both methods receive the same $\vz$: the comparator fits an MLP directly on these features, whereas AdaKerNet jointly learns the kernel feature map $\vphi_\psi$ and the nonlinear prediction head $h_\omega$. This comparison evaluates the complete second stage, including its architecture and joint objective, beyond the contribution of the first block.

AdaKerNet achieves higher mean test $R^2$ in 54 of 60 configurations, with reported relative $R^2$ gains reaching $25.2\%$. On AMAZON-FASHION for example, with LLaVA-1.5 and $N=100$, the score increases from $0.0963$ for MLP on $\vz$ to $0.1206$ for AdaKerNet. On the same benchmark with BLIP-2 and $N=100$, it increases from $0.1810$ to $0.2077$, while on QUECHUA-VALENCE at $N=100$, it increases from $0.0511$ to $0.0620$. These examples show that joint kernel and predictor learning can improve prediction performance even after the first block has already adapted the frozen representations.

In the remaining six configurations, AdaKerNet's shortfall is at most $0.0040$ in mean $R^2$. The exceptions occur on PAD-UFES-20-AGE with BLIP-2 at $N=100$ and Qwen2.5-VL at $N=200$; PAD-UFES-20-MOLE-LESION with LLaVA-1.5 at $N=500$ and $N=1000$, and Qwen2.5-VL at $N=500$; and SPEECHOCEAN762-PROSODIC at $N=200$. 

Table~\ref{tab:deepmlpz_vs_adakernet} further compares AdaKerNet with an architecture-matched deeper MLP on $\vz$, obtained by removing the kernel-reconstruction loss ($\mathcal{L}_{\mathrm{ker}}=0$). AdaKerNet achieves higher mean test $R^2$ in 53 of 60 configurations, with reported relative gains reaching $30.2\%$ and shortfalls of at most $0.0059$, supporting the contribution of kernel-reconstruction regularization beyond network depth alone.

\paragraph{Learned kernel versus reference kernel using a KRR head.}
Tables~6--11 evaluate whether the adaptively deformed learned kernel improves prediction compared to the reference kernel. We replace AdaKerNet's nonlinear prediction head with a KRR head using the learned kernel
\[
\widehat{k}_\psi(\vz,\vz')
=\vphi_\psi(\vz)^{\top}\vphi_\psi(\vz'),
\]
and compare it with KRR using the reference kernel $k_0$. Within each comparison, both predictors operate on the same fixed Lipschitz-controlled features $\vz$ and select the ridge-parameter  based on the available labeled points. The learned map is obtained through AdaKerNet's joint training; the subsequent KRR evaluation as a predominant kernelized method tests whether the resulting similarities are useful.

Across the four reference-kernel choices (RBF and Mat\'ern-$1/2$, $3/2$, and $5/2$) the reported relative gains favor $\widehat{k}_\psi$ in 217 of the 240 configurations in Tables~6--11. For example, on QUECHUA-VALENCE with the Mat\'ern-$3/2$ reference kernel at $N=100$, using $\widehat{k}_\psi$ increases $R^2$ from $0.0184$ to $0.0596$. On AMAZON-FASHION with BLIP-2 and an RBF reference kernel at $N=100$, it increases $R^2$ from $0.0837$ to $0.1742$. 

These results indicate that the benefit of joint training extends beyond the nonlinear head used by AdaKerNet: the deformed kernel also supports more accurate prediction with KRR in most evaluated settings. The reconstruction loss encourages the learned similarities to retain the reference structure, while supervised prediction allows task-relevant deviations. 

\subsection{Q1: Sensitivity to the reference-kernel choice}
\label{sec:supp_reference_choice}

Tables~12--13 examine whether AdaKerNet's advantage over baselines B-I--B-V depends on a particular reference kernel. We evaluate RBF and Mat\'ern-$1/2$, $3/2$, and $5/2$ reference kernels on PAD-UFES-20-AGE with BLIP-2, LLaVA-1.5, and Qwen2.5-VL, and on SPEECHOCEAN762-FLUENCY with Gemini Embedding~2. Five label budgets yield 80 benchmark--MLLM encoder--kernel--budget configurations. 

AdaKerNet outperforms B-II--B-V in all 80 configurations and the direct MLP baseline B-I in 78. All four reference-kernel choices outperform every baseline on PAD-UFES-20-AGE. The two exceptions occur on SPEECHOCEAN762-FLUENCY at $N=100$: AdaKerNet obtains $R^2=0.2033$ with RBF and $0.2056$ with Mat\'ern-$5/2$, compared with $0.2092$ for B-I. At the same budget, Mat\'ern-$1/2$ and Mat\'ern-$3/2$ yield $0.2168$ and $0.2130$, respectively, exceeding B-I. At every larger budget, all four choices outperform all five baselines on this benchmark.

No single reference kernel is best throughout. For instance, on SPEECHOCEAN762-FLUENCY, Mat\'ern-$1/2$ gives the highest AdaKerNet score at $N=100$ and $N=200$, while RBF gives the highest score at the three larger budgets. Overall, the results show that AdaKerNet's advantage is not restricted to one reference-kernel choice.

\subsection{Q2: Performance in the non-sparse label regime}
\label{sec:supp_larger_budgets}

Tables~14--19 evaluate non-sparse budgets above 1,000 labels, ranging from $N=1043$ to $N=3486$. These experiments cover all six benchmarks and include settings that use the full available training pool. AdaKerNet retains the benchmark-specific reference kernel used in the main comparison, while the five baselines operate directly on the frozen representations. There are 16 benchmark--encoder--budget configurations in total.

AdaKerNet outperforms all five baselines simultaneously in 12 of 16 configurations. It leads on all three MLLM encoder choices for PAD-UFES-20-MOLE-LESION, all six configurations for AMAZON-FASHION, and SPEECHOCEAN762-FLUENCY at $N=2000$. It also leads on PAD-UFES-20-AGE with LLaVA-1.5 and Qwen2.5-VL. For example, on AMAZON-FASHION with LLaVA-1.5 at $N=3486$, AdaKerNet achieves $R^2=0.5019$, compared with $0.4495$ for the strongest baseline, KRR. On PAD-UFES-20-AGE with Qwen2.5-VL at $N=1612$, it achieves $0.4106$, compared with $0.3519$ for KRR.

Overall, AdaKerNet's empirical advantage is not confined to sparse supervision, nor is it universal at larger budgets.

\subsection{Q3: Effect of the loss weight $\lambda$}
\label{sec:supp_lambda}

Table~20 investigates the balance between kernel reconstruction and supervised prediction in
\[
\mathcal{L}=\mathcal{L}_{\mathrm{ker}}
+\lambda\mathcal{L}_{\mathrm{pred}}.
\]
The parameter $\lambda$ balances reference-kernel reconstruction fidelity and task-specific adaptation: excessively small values may provide insufficient predictive supervision, whereas excessively large values weaken the relative influence of kernel reconstruction. As shown in Table~20 of the supplementary material for QUECHUA-VALENCE, $\lambda$ selected from $\{0.1,1,10\}$ yields the best result at every label budget and outperforms the extreme values in 10 of 12 kernel--budget configurations. This indicates that intermediate values of $\lambda$ generally outperform extremely low or high values, underscoring the importance of balancing kernel reconstruction and task-specific adaptation.

\subsection{Alternative kernel for the KRR baseline}
\label{sec:supp_alternative_krr}

Table~21 tests whether AdaKerNet's advantage over B-IV is specific to the RBF kernel used by that baseline in the main comparison. We replace the baseline's kernel with Mat\'ern-$1/2$ while retaining KRR on the raw representations $\ve$. AdaKerNet continues to use the benchmark-specific reference kernel from the main results. This experiment covers all six benchmarks, 12 benchmark--MLLM encoder pairs, and five label budgets, yielding 60 configurations.

AdaKerNet achieves higher mean test $R^2$ in all 60 configurations. Its relative reduction in test MSE ranges from $1.6\%$ to $41.5\%$. The largest reduction occurs on SPEECHOCEAN762-FLUENCY at $N=100$, where AdaKerNet achieves $R^2=0.2033$ compared with $-0.3622$ for Mat\'ern-$1/2$ KRR. The advantage also persists where the baseline predicts well: on AMAZON-FASHION with BLIP-2 at $N=1000$, AdaKerNet obtains $0.4888$ compared with $0.3747$, corresponding to an $18.2\%$ MSE reduction.

These results show that AdaKerNet's advantage over direct KRR persists when the baseline uses this alternative kernel.

\subsection{Comparison with spectral kernel learning}
\label{sec:ablation_trf}

To further assess the benefits of AdaKerNet's task-adaptive
kernel deformation stage, we compare it with Tuned Random
Features (TRF)~\citep{shilton2022trf}. TRF adapts a
translation-invariant kernel by learning spectral weights
over frequencies sampled from a reference density. AdaKerNet
instead learns a nonlinear feature map whose inner products
induce a positive-semidefinite kernel, jointly optimizing
kernel reconstruction and supervised neural prediction.
The resulting kernel need not remain translation-invariant,
allowing more flexible adaptation of sample similarities
to the downstream task.

Tuned Random Features
\citep{shilton2022trf} on the Lipschitz-controlled features, were implemented drawing
$D=256$ frequencies once from the spectral density of the benchmark's
reference kernel and learning a per-frequency density ratio $u \ge 1$
jointly with the read-out weights by Adam, with $u$ clipped after every
step as the original prescribes. 

Both methods operate on the same first-block features $\vz$
and use the same benchmark-specific reference kernel.
Thus, this comparison evaluates their respective decoding
strategies while keeping the input features and reference
kernel fixed. Table~\ref{tab:trfz_vs_adakernet} reports results
across six benchmark-- MLLM encoder combinations and five label
budgets, yielding 30 configurations.

AdaKerNet achieves higher mean test $R^2$ in 26 of the
30 configurations. Its largest relative gains occur at
$N=100$ across all six benchmark--encoder combinations,
ranging from $17.4\%$ to $137.7\%$. The latter corresponds
to an increase from $0.0261$ to $0.0620$ on
QUECHUA-VALENCE. AdaKerNet also outperforms TRF at every
evaluated label budget on PAD-UFES-20-MOLE-LESION with
LLaVA and on both SPEECHOCEAN762 tasks. The four exceptions
occur at larger label budgets: PAD-UFES-20-MOLE-LESION
with BLIP-2 at $N=500$ and $N=1043$, the same task with
Qwen at $N=1043$, and QUECHUA-VALENCE at $N=1000$.
These shortfalls do not exceed $0.0099$ in absolute $R^2$.

Overall, these results support the predictive benefits
of AdaKerNet's joint kernel deformation and nonlinear
neural prediction framework relative to spectral kernel
learning, particularly at the smallest label budget.
Because the methods also differ in their prediction
architectures and training objectives, the comparison
assesses the complete decoding strategies rather than
isolating kernel flexibility alone.

\newpage
\bibliography{iclr2027_conference}
\bibliographystyle{iclr2027_conference}

\end{document}